%% file: root.tex
\documentclass[letterpaper, 10 pt, journal, twoside]{IEEEtran}

\bstctlcite{bstctl:etal, bstctl:nodash, bstctl:simpurl}

\usepackage{graphicx}
\usepackage{amsmath}
\usepackage{times}
\usepackage{graphicx}
\usepackage{amssymb}
\usepackage{gensymb}
\usepackage{textcomp}
\usepackage{multirow}
\usepackage{booktabs}
\usepackage{flushend}
\usepackage{multicol}
\usepackage{color}
\usepackage{makecell}
\usepackage{bbding}
\usepackage{array}
\usepackage{pifont}
\usepackage{colortbl}
\usepackage{cite}
\usepackage{bm}
\usepackage{mfirstuc}
\usepackage[dvipsnames]{xcolor}

\usepackage[colorlinks,pagebackref=false,citecolor=blue,bookmarks=false,hypertexnames=true]{hyperref}
\usepackage[figureposition=bottom,tableposition=top]{caption}
\ifx00 
\newcommand{\AD}[1]{\textcolor{blue}{[Arindam: #1]}}
\else 
\newcommand{\AD}[1]{\textcolor{blue}{}}
\fi

\begin{document}

\title{Fisheye Camera and Ultrasonic Sensor Fusion For Near-Field Obstacle Perception in Bird's-Eye-View}

\author{
Arindam Das,
Sudarshan Paul,
Niko Scholz, 
Akhilesh Kumar Malviya,
Ganesh Sistu, \\
Ujjwal Bhattacharya~\IEEEmembership{Senior Member,~IEEE}, and
Ciarán Eising~\IEEEmembership{Senior Member,~IEEE}

\thanks{A. Das, G. Sistu, and C. Eising are with the Department of Electronic and Computer Engineering, University of Limerick, Limerick, V94 T9PX Ireland. A. Das, G. Sistu, and C. Eising are also with the Data Driven Computer Engineering Research Group, University of Limerick, Limerick, V94 T9PX Ireland.}
\thanks{A. Das, S. Paul, and A. K. Malviya are with the Department of Driving Software and Systems, Valeo India, Chennai 600130, India}
\thanks{N. Scholz is with Valeo Driving Assistance Research, Hummendorfer Str. 74, 96317 Kronach, Germany.}
\thanks{G. Sistu is with Valeo Vision Systems, Dunmore Road, Tuam, Co. Galway, H54 Y276 Ireland.}
\thanks{U. Bhattacharya is with the Computer Vision and Pattern Recognition (CVPR) Unit, Indian Statistical Institute, Kolkata 700108, India.}
}


\maketitle

\bstctlcite{IEEEexample:BSTcontrol}

\newcommand{\etal}{\textit{et al.}}
\newcommand{\xmark}{\ding{55}}

\input{include/abstract.tex}

\begin{IEEEkeywords}
Multimodal Learning, Multimodal Feature Fusion, Obstacle Perception, Bird's-Eye-View.
\end{IEEEkeywords}

\IEEEpeerreviewmaketitle

\input{include/introduction.tex}

\input{include/related_work.tex}

\input{include/sensors.tex}

\input{include/dataset.tex}

\input{include/architecture.tex}

\input{include/results.tex}

\input{include/conclusions.tex}



\bibliographystyle{IEEEtran}
\bibliography{IEEEfull}

\input{include/bio}


\end{document}

%% file: include/abstract.tex
\begin{abstract}
Accurate obstacle identification represents a fundamental challenge within the scope of near-field perception for autonomous driving. Conventionally, fisheye cameras are frequently employed for comprehensive surround-view perception, including rear-view obstacle localization. However, the performance of such cameras can significantly deteriorate in low-light conditions, during nighttime, or when subjected to intense sun glare. Conversely, cost-effective sensors like ultrasonic sensors remain largely unaffected under these conditions. Therefore, we present, to our knowledge, the first end-to-end multimodal fusion model tailored for efficient obstacle perception in a bird's-eye-view (BEV) perspective, utilizing fisheye cameras and ultrasonic sensors. Initially, ResNeXt-50 is employed as a set of unimodal encoders to extract features specific to each modality. Subsequently, the feature space associated with the visible spectrum undergoes transformation into BEV. The fusion of these two modalities is facilitated via concatenation. At the same time, the ultrasonic spectrum-based unimodal feature maps pass through content-aware dilated convolution, applied to mitigate the sensor misalignment between two sensors in the fused feature space. Finally, the fused features are utilized by a two-stage semantic occupancy decoder to generate grid-wise predictions for precise obstacle perception. We conduct a systematic investigation to determine the optimal strategy for multimodal fusion of both sensors. We provide insights into our dataset creation procedures, annotation guidelines, and perform a thorough data analysis to ensure adequate coverage of all scenarios. When applied to our dataset, the experimental results underscore the robustness and effectiveness of our proposed multimodal fusion approach. A short video highlighting an overview of this work along with its more qualitative results can be seen at \url{https://youtu.be/JmSLBBL9Ruo}.

\end{abstract}

%% file: include/introduction.tex
\section{Introduction}
\begin{figure}
    \captionsetup{singlelinecheck=false, font=small}
    \centering
    \includegraphics[width=0.49\textwidth]{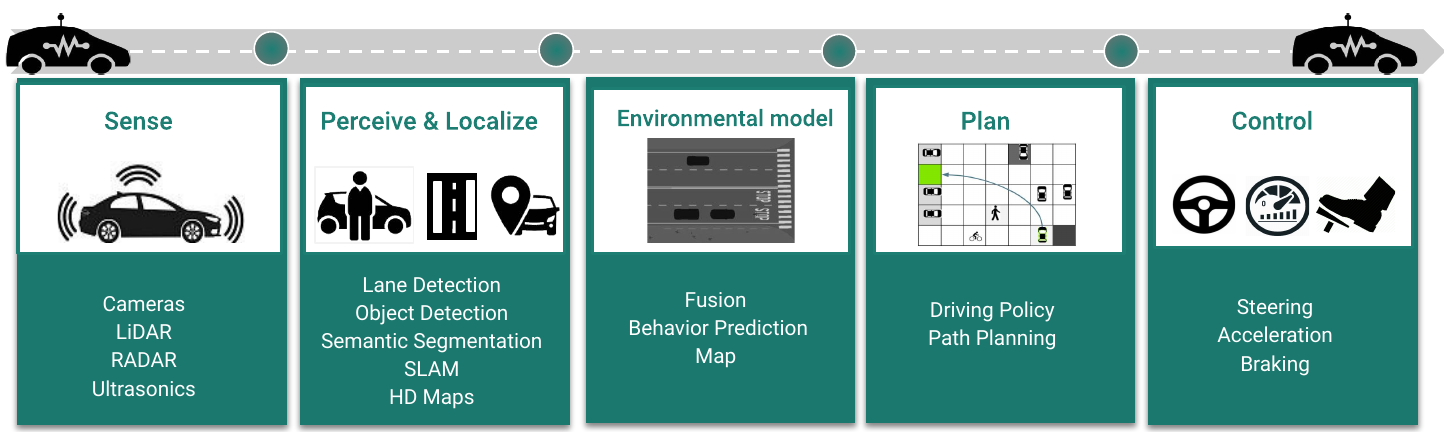}\\
    \caption{ \textbf{A standard autonomous driving pipeline \cite{uricar2019yes}}
    }
    \vspace{-5mm}
    
    \label{fig:adas_pipeline}
\end{figure}

Autonomous driving is steadily becoming a standard feature in contemporary vehicles \cite{yurtsever2020survey}. However, achieving full autonomy in driving remains a formidable challenge, prompting active research to tackle the intricate technical issues involved. Figure \ref{fig:adas_pipeline} provides a typical end-to-end pipeline containing essential components of a standard autonomous driving system \cite{xu2019design}. The initial stage involves perception, wherein a suite of sensors such as cameras, radar, ultrasonics, and LiDAR come into play. During the perception stage, the system identifies semantic elements like lanes and vehicles, as well as geometric entities such as free space and general obstacles. Subsequently, these findings are combined into an abstract representation, typically manifesting as a 2D or 3D map depicting objects about the ego vehicle. To make informed driving decisions, a driving policy algorithm leverages this map to determine the most suitable trajectory for the vehicle's maneuvering. While the traditional approach involves the independent design of these modules, ongoing endeavors are also exploring the potential of end-to-end learning methodologies.

As per a report in \cite{singh2015critical} released by the National Highway Traffic Safety Administration, human errors, including instances of crashing vehicles into obstacles, contribute to as much as 94\% of road accidents. In this context, the development of autonomous driving, while currently not a mature technology, through driver assistance systems has the potential to prevent accidents, decrease emissions, provide transportation for individuals with limited mobility, and alleviate driving-related stress \cite{crayton2017autonomous}. Cameras take precedence as the primary sensor \cite{li2023emergent} due to their alignment with transportation infrastructure designed for human optical perception. Moreover, they offer a cost-effective means of capturing rich semantic and geometric information. In particular, surround-view cameras are used for short-range perceptions \cite{HEIMBERGER201788} such as parking, low-speed maneuvering, emergency braking systems, etc.

In this paper, we focus on obstacle perception tasks where obstacles are not in any predefined categories. This consideration makes our work different than general fully supervised object detection methods. Fisheye cameras play an important role in our application, however, camera sensors introduce several critical challenges when operating in real-world scenarios:

\textbf{Inadequate Low-Light Performance:} Cameras struggle in scenarios with low light levels, typically below 2.5 lux \cite{thesis}, depending on the exact sensor and task. This limited illumination severely degrades the quality of semantic representation of obstacles, impacting related applications such as pedestrian pose estimation \cite{cao2017realtime} and activity recognition \cite{minguez2018pedestrian}.

\textbf{Exposure to Environmental Elements:} Automotive surround-view cameras are commonly mounted externally, exposing the camera lens to the risk of contamination from substances like sand, mud, dirt, snow, or grass \cite{das2020tiledsoilingnet}.

\textbf{Sun Glare Interference:} The presence of sun glare can result in overexposed regions on the camera lens \cite{yahiaoui2020let}, hindering the efficiency of downstream vision-based algorithms for obstacle perception.

\textbf{Densely Cast Shadows:} Dense shadows pose challenges \cite{le2021physics}, particularly for algorithms operating at the pixel level, including tasks like semantic segmentation \cite{minaee2021image} and instance segmentation \cite{cai2019cascade}.

These challenges highlight the need for the inclusion of additional sensors such as infrared for pedestrian detection \cite{liu2019ptb, das2023revisiting}. To achieve a dependable near-field perception \cite{eising2021near}, the focus of this paper is on identifying obstacles from a bird's-eye view. Ultrasonic sensors are a suitable technology to combine with a fisheye camera. They offer cost-effective short to moderate-range object detection, and low power consumption, are insensitive to the color or material of objects, resistant to ambient light, making them valuable for autonomous driving. 

This paper introduces an end-to-end CNN-based fusion model, designed for obstacle perception in a bird's-eye view by leveraging both fisheye and ultrasonic sensor data. By employing a preprocessing step, the raw echo amplitudes generated by the ultrasonic sensor are transformed into images that can be fed as inputs to a Convolutional Neural Network (CNN). We use a standard encoder to extract unimodal features from each sensor and later fuse them to produce modality-agnostic features. Finally, multimodal features are transformed into a bird's eye view, followed by a semantic segmentation decoder that performs pixel-level classification for an obstacle.

\textbf{Summary of the key contributions of this work:}

\begin{enumerate}
    \item We introduce a novel multisensor deep network tailored for near-field obstacle perception in a bird's-eye view. The proposed network combines a fisheye camera and an ultrasonic sensor system, marking the first known effort in this direction.
    \item We establish the strategy for creating a multisensor dataset comprising fisheye and ultrasonic data. We define annotation rules and present relevant data statistics that are essential for constructing a multimodal model suitable for similar applications.
    \item We describe an implementation of an end-to-end trainable network achieving very high accuracy. Additionally, we propose refactoring our proposal to support the same feature with unimodal inputs to carry out an in-depth analysis of the benefits of having a multimodal-based solution.
    \item We conduct comprehensive ablation studies involving various proposed network components, different feature fusion techniques,  diverse augmentation methods, and various loss functions.
\end{enumerate}

%% file: include/related_work.tex
\section{Related Work}
In this section, we discuss a detailed overview of the existing literature related to obstacle perception that includes visible and ultrasonic sensor technologies. Further, we provide a survey of the same function built using other automotive sensors. Next, the discussion goes into the broader domain of sensor fusion in perspective view for obstacle perception and then bird's-eye-view (BEV). It is also to be noted that only a very few works are found in the literature that deal with obstacles or unclassified object perception.

\subsection{Visible spectrum-based obstacle perception}
In contrast to 3D sensing, Levi \etal \cite{levi2015stixelnet} employed a single color camera, reducing the problem to column-wise regression that was solved using a deep CNN and a novel loss function based on semi-discrete obstacle position probability representation for network training. Ohgushi \etal \cite{ohgushi2020road} presented a road obstacle detection method by adding an autoencoder with semantic segmentation trained solely on normal road scenes. This method takes a color image from an in-vehicle camera as input and generates a resynthesized image using a semantic image generator as the encoder and a photographic image generator as the decoder. Then existing methods used external datasets or complex training for unexpected objects, requiring substantial effort or time, however, the authors in \cite{jung2021standardized} proposed a simple solution to standardize prediction scores, enhancing the identification of unexpected objects in urban scene segmentation. 
Traditional column-based obstacle detection identifies 3D obstacles but lacks object classification, segmentation, and motion prediction. Garnett \etal \cite{garnett2017real} introduced a single efficient deep convolutional network that combines these functions for real-time performance. Training of their proposal utilized both manual and automatically generated Lidar annotations. Vojir \etal \cite{vojir2021road} framed the issue of anomaly detection assuming that unfamiliar objects cannot be learned. The authors introduced a reconstruction module for semantic segmentation networks, using road surface reconstruction as a strong anomaly indicator and combining it with structural similarity error for per-pixel anomaly scores.


\begin{figure}
    \captionsetup{singlelinecheck=false, font=small}
    \centering
    \includegraphics[width=1.68in, height=1.4in]{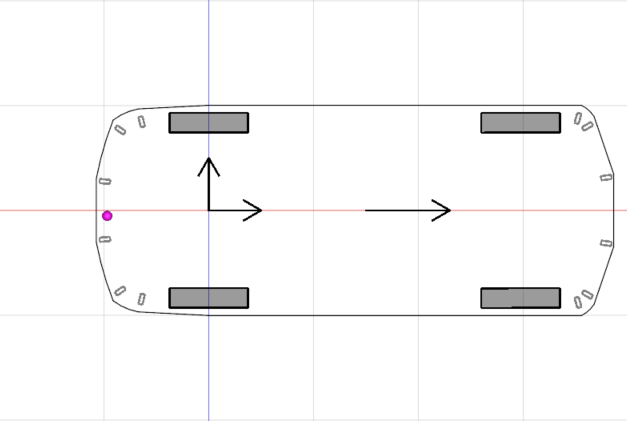}
    \includegraphics[width=1.68in, height=1.4in]{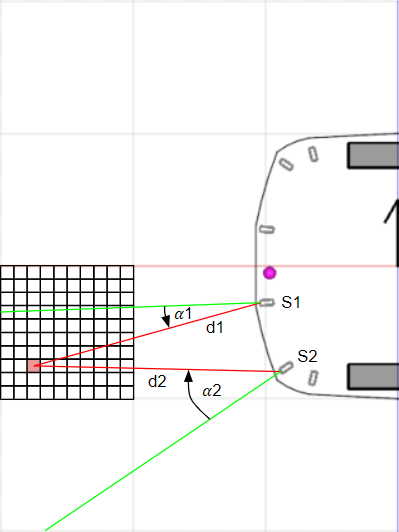}\\
    (a)  \hspace{35mm} (b)
    \caption{
    \textbf{(a) Mounting positions for the ultrasonic sensors and the fisheye camera.} The twelve ultrasonic sensors are shown as grey boxes on the outline of the car, and the rear fisheye camera is shown as a pink dot at the rear of the car, both symbols are not to scale with the actual sensor. The coordinate system corresponds to ISO 8855. \textbf{(b) Schematic of an ultrasonic grid map filling step for one exemplary grid cell and one exemplary signalway.} The grid is not to scale. The signal was emitted by Sensor $S1$ and is received by Sensor $S2$. To get the echo amplitude value at the highlighted grid cell, the distances $d1$ and $d2$ as well as the angles $\alpha1$ and $\alpha2$ used for amplitude attenuation are determined.
    }
    
    \vspace{-8mm}

    \label{fig:sensor_positions}
\end{figure}

\subsection{Ultrasonic sensor-based obstacle perception}

Ultrasonic sensors provide echolocation for perceiving the environment, supporting advanced driver assistance systems (ADAS). These systems detect obstacles \cite{kodera1984rear} and inform the driver via visual or auditory signals through human-machine interfaces (HMI). Figure \ref{fig:sensor_positions} (a) shows a general setup for an array of ultrasonic sensor systems to detect unclassified obstacles. The authors in \cite{mohamed2019convolution} introduced a machine learning-based technique to reduce noise in signals from automotive-grade active ultrasonic sensors. The reported method employed a Convolutional Neural Network (CNN) and was compared against state-of-the-art noise suppression techniques. The method was evaluated using both simulated scenarios and real sensor measurements, employing various metrics to assess signal quality and the effectiveness of different noise suppression approaches.

\subsection{Other sensor-based obstacle perception}
Hancock \cite{hancock1997high} employed laser reflectance and stereo vision to identify small road obstacles located at significant distances. Likewise, William \etal \cite{williamson1998detection} applied a multibaseline stereo technique to detect small road obstacles, which were approximately 14 cm high and located at distances exceeding 100 meters. 
In another stereo vision-based approach \cite{labayrade2002real}, non-flat road geometry was estimated for obstacle detection using the ``v-disparity" image. Subaru Eyesight \cite{SUBARU} stands out as a stereo-vision-based system renowned for its robust capability in detecting sizable road obstacles. Additionally, Mobileye \cite{Mobileye} is a commercially accessible system known for its robustness in identifying large obstacles when situated in close proximity, relying solely on a monocular camera.

\subsection{Multimodal obstacle perception}
The method in \cite{manduchi2005obstacle} introduced sensor algorithms for cross-country autonomous navigation, utilizing a combination of a color stereo camera and a single-axis LADAR (LAser Detection And Ranging). The author proposed an obstacle detection method based on stereo range measurements, without relying on scene structure assumptions. Additionally, a color-based system for classifying obstacles into terrain categories was presented along with an algorithm for analyzing LADAR data to distinguish grass from hidden obstacles like tree trunks or rocks. The authors in \cite{bertozzi2008obstacle} introduced a system designed to detect and categorize road obstacles, such as pedestrians and vehicles. The reported system integrated data from various sensors, including a camera, radar, and inertial sensor. The primary role of the camera was to enhance the precision of radar-detected vehicle boundaries and eliminate potential false positives. Simultaneously, a symmetry-based pedestrian detection algorithm was employed, and its outcomes were combined with a set of regions of interest obtained through a Motion Stereo technique. Many obstacle-detection methods are constrained by specific scenarios and conditions. In contrast, the method proposed in \cite{shinzato2014road} highlighted a robust sensor fusion approach capable of detecting obstacles across a broad range of scenarios with minimal parameter requirements. The method relied on analyzing the spatial relationships within perspective images from a single camera and a 3D LiDAR sensor. Ramos \etal \cite{ramos2017detecting} 
proposed a deep learning-based obstacle detection system that combines appearance and context cues, employing a modified fully convolutional network to semantically label pixels as free space, on-road unexpected obstacles, or background.  
On the other hand, camera and LiDAR point cloud data were used in \cite{bogdoll2022multimodal} to detect road anomalies by sequentially employing latest state-of-the-art detection models.

\subsection{Sensor fusion in BEV}
Liang \etal \cite{liang2022bevfusion} introduced BEVFusion, a novel fusion framework that operates independently of LiDAR data, overcoming this limitation in previous methods. Current methods use point-level fusion, adding camera features to the LiDAR point cloud. However, this approach sacrifices the semantic information in-camera features, making it less effective for tasks like 3D scene segmentation. In contrast, the method in \cite{liu2023bevfusion} introduced BEVFusion, an efficient multi-sensor fusion framework. BEVFusion unifies multimodal features within a shared BEV representation space, preserving both geometric and semantic information. Existing approaches often overlook depth information when converting 2D data to 3D, leading to unreliable fusion of 2D semantics with 3D points. 
Jiao et al. \cite{jiao2023msmdfusion} addressed this limitation and introduced an improved framework for better utilization of the depth information and achieving fine-grained cross-modal interactions between LiDAR and the camera. In the method proposed in this paper, we overcome this limitation by completing the fusion in 2D space, not with 3D points.
In another recent work, Man et al. \cite{man2023bev} presented BEVGuide, a representation learning framework,  
accommodates input from a diverse sensor array, such as cameras, LiDAR, and radar.
The method introduced a BEV-guided multi-sensor attention block that uses BEV embeddings to learn representations from sensor-specific features.

In this paper, we make the first attempt towards obstacle perception in BEV using a fisheye camera and an ultrasonic sensor system where obstacles are not in any well-defined categories, rather they are unclassified objects.

%% file: include/sensors.tex
\section{Automotive Sensors for near-field sensing}
\begin{figure}
    \captionsetup{singlelinecheck=false, font=small}
    \centering
    \includegraphics[width=0.42\textwidth]{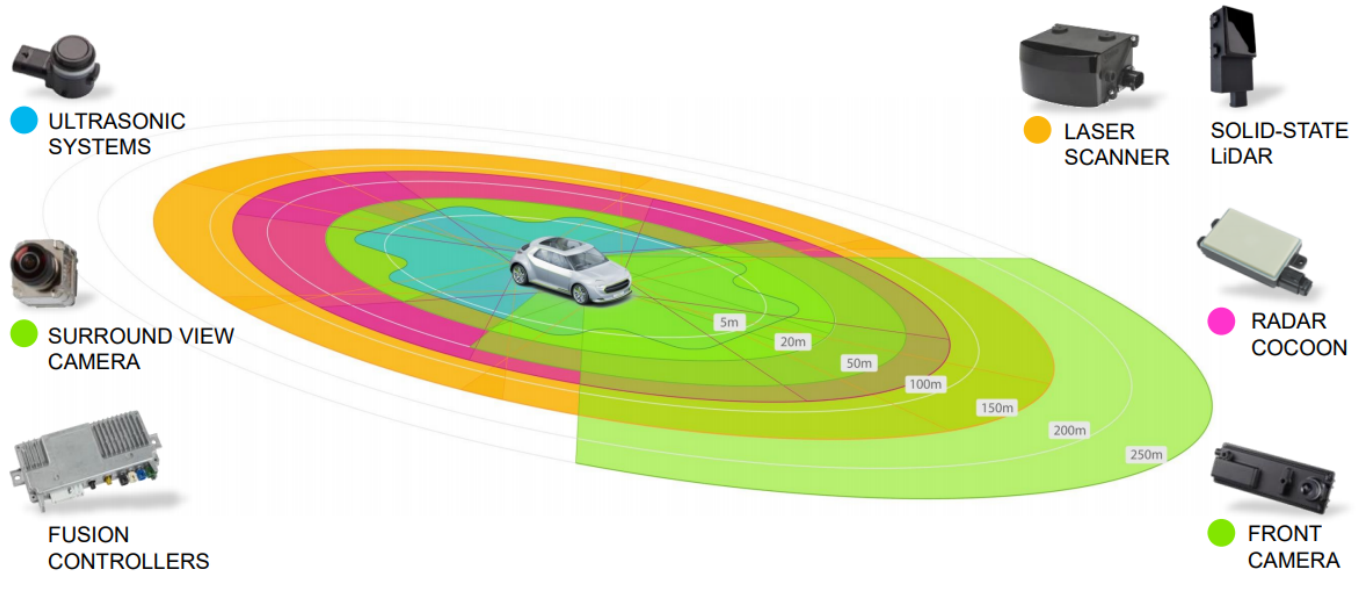}\\
    \caption{ \textbf{Various automotive sensors \cite{eising2021near} used in a typical perception stack} in either unimodal or multimodal settings.
    }
    \vspace{-4mm}  
    \label{fig:automotive_sensors}
\end{figure}

Automotive sensors such as ultrasonic sensors, surround-view cameras, radars, LiDAR, and long-range front cameras as shown in Figure \ref{fig:automotive_sensors}, are strategically placed around the vehicle to infer objects and obstacles in the immediate vicinity. They are crucial for tasks like parking assistance, blind-spot monitoring, and collision avoidance in low-speed and complex urban environments. As autonomous driving continues to evolve, accurate near-field sensors are crucial for ensuring the safety and success of self-driving vehicles in challenging real-world scenarios. In this section, first, we discuss the primary sensors used in this work - the fisheye camera and the ultrasonic sensor. We also provide high-level details of other automotive sensors such as radar, sonar, front camera, and LiDAR.

\subsection{Fisheye camera}

\begin{figure}
    \captionsetup{singlelinecheck=false, font=small}
    \centering
    \includegraphics[width=0.488\textwidth]{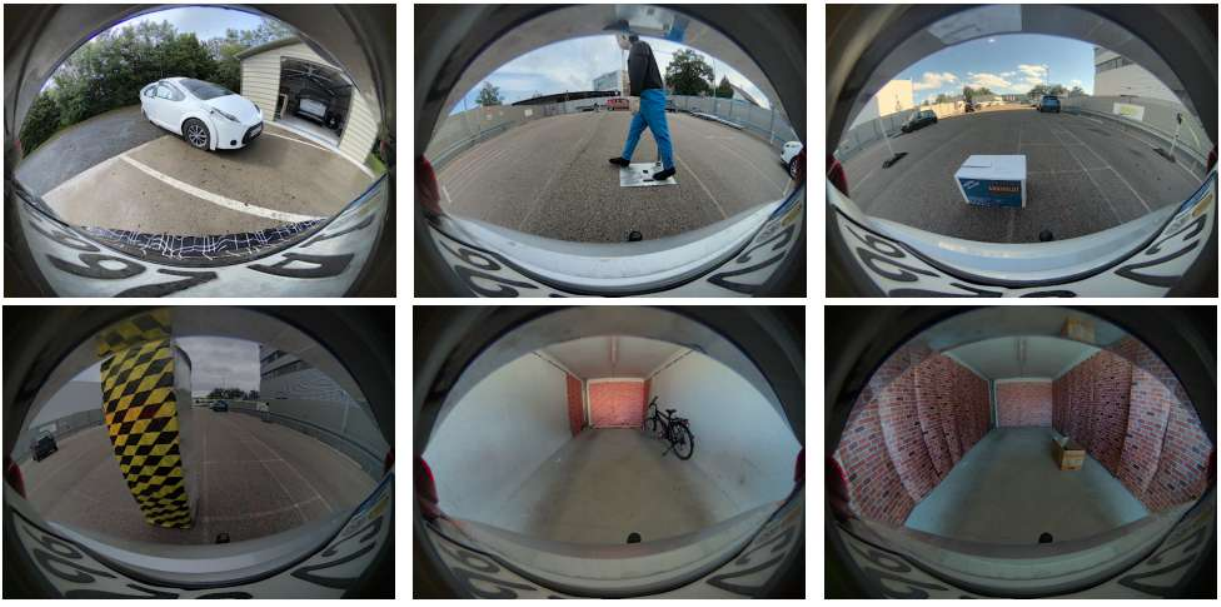}\\
    \caption{\textbf{Different types of obstacles commonly appear in rear-view}. Top: vehicle, pedestrian, carton. Bottom: pillar, cycle, wooden box.}
    \vspace{-5mm}
    \label{fig:different_obstacles}
\end{figure}

Fisheye cameras are widely employed for close-range sensing in automotive perception applications. Using four fisheye cameras positioned on each side of the vehicle can achieve 360° coverage, encompassing the entire near-field sensing area. This setup is suited to applications like automated parking, traffic jam assistance, urban driving, and rear obstacle detection. Fisheye lenses are employed to expand the field of view of cameras to 180° or beyond, offering advantages in minimizing the number of cameras to observe the vehicle surroundings. However, any perception algorithm needs to account for the considerable fisheye distortion inherent in such camera systems, as illustrated in Figure \ref{fig:different_obstacles}. The reader is referred to \cite{kumar2023surround} \cite{eising2021near} for a more complete overview of fisheye cameras, and their uses in surround-view systems.

\subsection{Ultrasonic sensor}
Ultrasonic sensors present a reliable, low-powered, and cost-effective means for detecting objects within limited to intermediate distances, thereby enhancing obstacle avoidance capabilities and streamlining parking maneuvers. Ultrasonic sensors operate in a non-contact mode, preserving performance regardless of object color or material, making them versatile for various road scenarios. 
Nevertheless, it is important to acknowledge that ultrasonic sensors do possess certain limitations, which include a limited detection range (less than 10 meters), and a constrained field of view (though with an array of sensors, this limitation can be resolved), vulnerability to adverse weather conditions, and challenges at high speed. To address these constraints effectively, ultrasonic sensors are commonly deployed \cite{valeo_uls} in close conjunction with other sensor technologies to establish a comprehensive and reliable perception system.

\subsection{Other related sensors}

\textbf{Short-Range Radar (SRR):} The coverage of a short-range radar \cite{stateczny2019empirical} sensor is up to 30 meters. For this reason, it is widely used for driver assistance applications for near-field maneuvering, such as parking assistance and blind-spot monitoring. However, this sensor is highly prone to high false positives when reflections from nearby objects like road signs and guardrails create additional signals that can be interpreted as real targets.

\textbf{Medium-Range Radar (MRR):} The standard coverage of this sensor \cite{stateczny2019empirical} is usually between 30 to 60 meters. It helps to detect other automotive actors such as vehicles from adjacent lanes. However, it is limited in terms of range at high speeds in highway driving scenarios. This radar also gets highly affected in adverse weather conditions.

\textbf{Long-Range Radar (LRR):} The coverage of this radar \cite{stateczny2019empirical} is up to 200 meters. Due to this long range, it helps to get distant objects detected early in highway scenes. However, it provides low-resolution data as compared to short-range radar.

\textbf{Long-Range Front Camera:} Based on the specification, the range of this sensor can be typically up to 200 meters or even more. Due to its high-resolution data, dense tasks can be performed with high accuracy. Front cameras are often more cost-effective than other sensors like LiDAR. Adverse weather conditions such as heavy rain, snow, or fog greatly hinder the performance of front cameras, thus affecting the overall efficiency of the perception stack. In addition, the field-of-view of front cameras is limited compared to LiDAR which can miss the object detection from all directions.

\textbf{Solid-state LiDAR:} The range of this sensor depends on the manufacturer, however it is mostly up to 200 meters. This sensor comes as more compact and robust than traditional mechanical LiDAR technology which makes it suitable for integration into various vehicle designs. It is also more cost-effective than mechanical LiDARs, making it a worthy sensor for mass production for automotive use cases. However, it may have a limited field of view and resolution as compared to traditional LiDARs.

%% file: include/dataset.tex
\section{Dataset Creation}
The ultrasonic data recorded in these scenes consists of echo amplitudes from twelve ultrasonic sensors mounted in the front and rear bumper of the car. Since our region of interest is the field of view of the rear fisheye camera, we are only using the data from the six rear-mounted ultrasonic sensors. See Figure \ref{fig:sensor_positions} (a) for exemplary mounting positions in the car for ultrasonic sensors and the rear fisheye camera. 

\subsection{Data capture}
\begin{figure}
    \captionsetup{singlelinecheck=false, font=small, belowskip=-6pt}
    \centering
    \includegraphics[angle=90,width=0.326\textwidth]{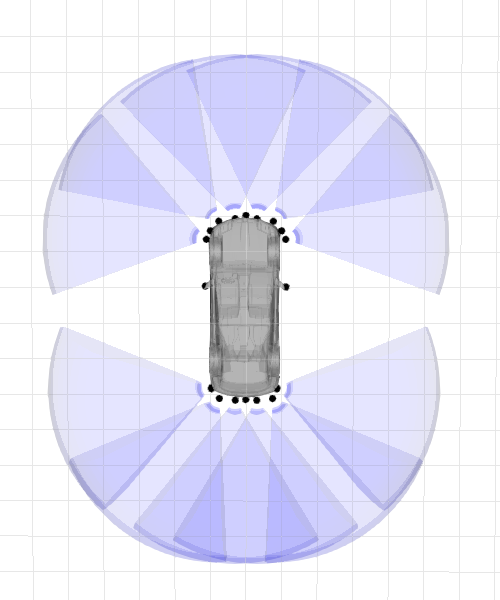}\\
    \caption{\textbf{Estimated Field of View of the ultrasonic sensor system.} The actual field of view also depends on the object being observed. Echos are strongest on the sensor axis and weaken for objects positioned off-axis. (Image courtesy of Nathaniel Arnest, DSW Kronach, nathaniel.arnest@valeo.com.)}
    \label{fig:uls_fov}
\vspace{-5mm}
\end{figure}

As an active sensor, an ultrasonic sensor will generate a signal and listen for echoes coming from objects in the surroundings. Sensors might also listen passively to echoes of signals generated by neighboring active sensors. The type of ultrasonic sensors \cite{valeo_uls} used for these recordings is not only able to report echo distances and peak properties like maximum amplitude and peak width but also the entire envelope of the received echo amplitude over the time it is listening for echoes.
The maximum distance it can receive echoes from therefore depends on listening time but also the speed of sound in air and thus directly on air temperature. It is limited by the dissipation of sound wave energy of the ultrasonic signal with travel distance, which degrades the signal-to-noise ratio in the received echo. In our recordings, the measurement sequence used limited the radial field of view of the sensors to less than 4.5m.
The horizontal field of view depends on a variety of factors, including the type and orientation of the object and the distance between the object and the sensor, but also the way the signal is received. so whether it is sent and received on the same active sensor, or emitted by one sensor but received by a neighboring passive sensor not emitting a signal in the same frequency in this measurement step. In general, the field of view will be larger for close objects and narrower for objects further away from the sensor. It is generally given as an opening angle of 130° to 70°. The positioning of the sensor on the car bumper will lead to substantial overlap in the field of view of the six ultrasonic sensors so that the entire near field of the fisheye camera will be in the field of view of multiple ultrasonic signals at any given time, see Figure \ref{fig:uls_fov}.

\begin{figure}
    \captionsetup{singlelinecheck=false, font=small, belowskip=-6pt}
    \centering
    \includegraphics[width=0.35\textwidth]{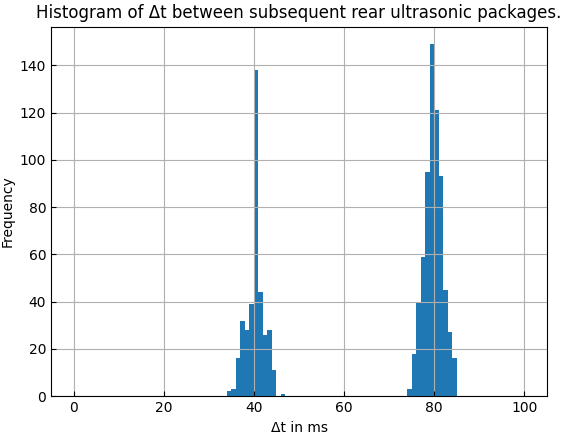}\\
    \caption{\textbf{Histogram of the difference in timestamps between subsequent ultrasonic packages in one exemplary scene.} The two major modes around 40 ms and 80 ms are a product of the measurement interval being longer than the update cycle of 40 ms.}
    \label{fig:timestamp_histogram}
\vspace{-1mm}
\end{figure}

However, there are two challenges we need to overcome to combine camera and ultrasonic data. First, the ultrasonic data is in a decidedly different domain than the fisheye images. For each measurement step, we receive eight arrays of the echo amplitude over time, where time equates to the traveled distance of the echo signal. Additionally, we have meta-information like which sensors are active or passive, where the sensors are located on the bumper, and how they are oriented relative to the car coordinate system.
Second, fisheye image data is recorded in 2-megapixel using a Surround View System (SVS) camera at a frequency of 30 frames per second, while ultrasonic measurements of the rear bumper system arrive at irregular intervals between 34 ms and 85 ms. The distribution has two major modes at 40 ms and 80 ms, as can be seen in Figure \ref{fig:timestamp_histogram}. The two modes are a result of the ultrasonic measurement frequency in conjunction with the 40 ms cycle of odometry packages. Ultrasonic data packages are sent on the same 40 ms schedule as the odometry packages, but since the ultrasonic measurement interval is around 66 ms, there are steps where no ultrasonic data is available to be sent.  

The two major tasks in the dataset preparation were: 1) to find and transform the ultrasonic and image data to a common domain to be able to fuse them in feature space, and 2) to synchronize the fisheye images with the ultrasonic data.
 
\subsection{Preprocessing}
As a common domain, we identified the bird’s-eye-view perspective, which is also the desired domain of the network output, and has been used for camera data previously in the literature. To transform the ultrasonic data, which is effectively an echo strength over distance measurement, to a BEV representation, we use a grid structure, anchored at the camera position in x and y coordinates but with a z-coordinate of zero, and extending six meters to the rear and to each side, to cover the entire field of view of the rear ultrasonic sensor system. The grid cell size is set to be 1 cm, which is smaller, but of the same magnitude, as the distance quantization of the echo amplitude array. 
For each grid cell and each signalway, the distance and the angle between the grid cell location and the emitting and receiving sensor are calculated, as shown in Figure \ref{fig:sensor_positions}(b). Both distances are added together and the corresponding amplitude value of the echo is interpolated linearly at this distance using the signalway amplitude array. Since the opening angle of the field-of-view of an ultrasonic sensor is limited, the angle $\alpha$ from the sensor axis to the grid cell is used to attenuate the amplitude towards high angles off-axis using a scaled beta distribution $f(\alpha; 2,2)$. The echo amplitude values per cell are summed across signalways and put in the grid. This way, we distribute the echo amplitudes of the eight different signalways over the grid and implicitly add the spatial meta-information of the sensor positions and orientations to the data.

To have both inputs available to the network at the same time, we decided to synchronize the two different data domains by exporting an ultrasonic BEV map for every image frame we have from the camera. This means using the same ultrasonic raw data for multiple images, as our fisheye camera has a higher frame rate than the ultrasonic sensor system. Since we also have odometry available every time we receive an ultrasonic frame, it is possible to compensate for the ego motion of the vehicle between the time the ultrasonic data was measured and the time the fisheye image was recorded.

\subsection{Dataset statistics}
\begin{figure}
    \captionsetup{singlelinecheck=false, font=small}
    \centering
    \includegraphics[width=1.68in, height=1.4in]{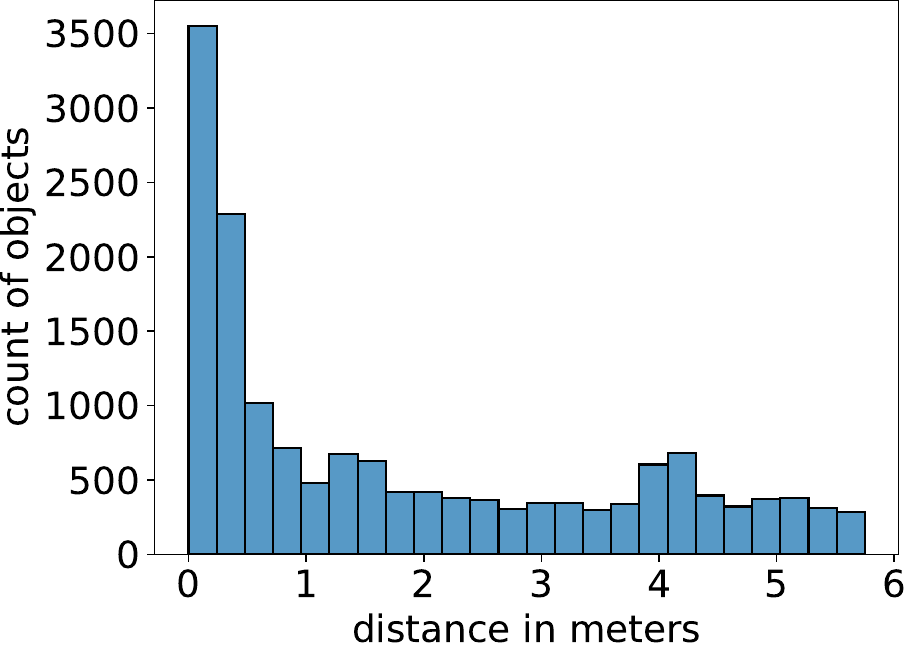}
    \includegraphics[width=1.68in, height=1.43in]{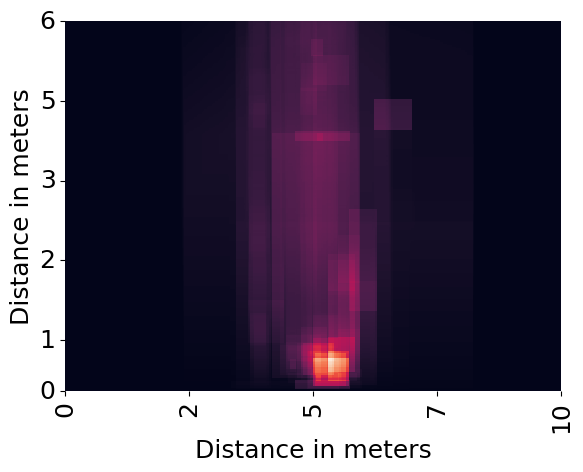}\\
    (a)  \hspace{35mm} (b)
    \caption{
    \textbf{(a) Object distance plot}. \textbf{(b) Heatmap plot of obstacles}.
    }
    
    \vspace{-5mm}

    \label{fig:data_stat}
\end{figure}

In this study, our dataset comprises a total of 35 scenes captured from the rear side of the ego vehicle, with each scene offering a rich array of data from both fisheye camera and ultrasonic sensors, along with their respective Bird's Eye View (BEV) semantic ground truth annotations. This comprehensive dataset was utilized to gain insights into the distribution of obstacle distances from the ego vehicle. Notably, a histogram plot in Figure \ref{fig:data_stat} (a) was generated, revealing that the majority of obstacles are concentrated within the 0-2 meters range, underscoring the prevalence of obstacles at short distances. Furthermore, through a heat map representation, Figure \ref{fig:data_stat} (b) illustrates the spatial distribution of obstacles as perceived in the BEV space. The heat map corroborates the observation that a substantial portion of the obstacles are near the ego vehicle. Intriguingly, it also highlights that the obstacles captured in the dataset span across various regions within the ego vehicle's field of view.

The x and y axes in the heat map provide valuable insights into the horizontal and vertical distances of obstacles from the position of the ego vehicle. A detailed analysis of the heat map reveals that obstacles are prominently concentrated at vertical distances of 0-2 meters from the ego vehicle, offering significant information about the distribution of obstacles at the rear side of the ego vehicle.

\subsection{Data split creation}
The dataset contains a comprehensive collection of 35 scenes, comprising a total of 15,928 frames. These frames include fisheye RGB images, ultrasonic BEV maps, and ground truth as segmentation masks in BEV, all focused on the task of semantic occupancy prediction of obstacles in BEV. The semantic annotations for obstacles have been executed from a top-view perspective. Approximately 8 distinct categories of obstacles are considered in this dataset (the obstacle categories are defined in Table \ref{tab:obstacle_type}). The dataset was systematically partitioned on a per-scene basis, with a deliberate allocation of 24 scenes exclusively for the training catalog. Among these, there were 19 outdoor scenes, while the remaining 5 scenes pertained to indoor obstacle scenarios. The validation set comprises a total of 3 scenes, with 2 scenes categorized as indoor scenes and the remaining 1 scene classified as outdoor scenes. On the other hand, the test set consists of 8 scenes, with 5 scenes falling into the outdoor category and 3 scenes designated as indoor scenes.

\begin{figure}
    \captionsetup{singlelinecheck=false, font=small, belowskip=-6pt}
    \centering
    \includegraphics[width=0.487\textwidth]{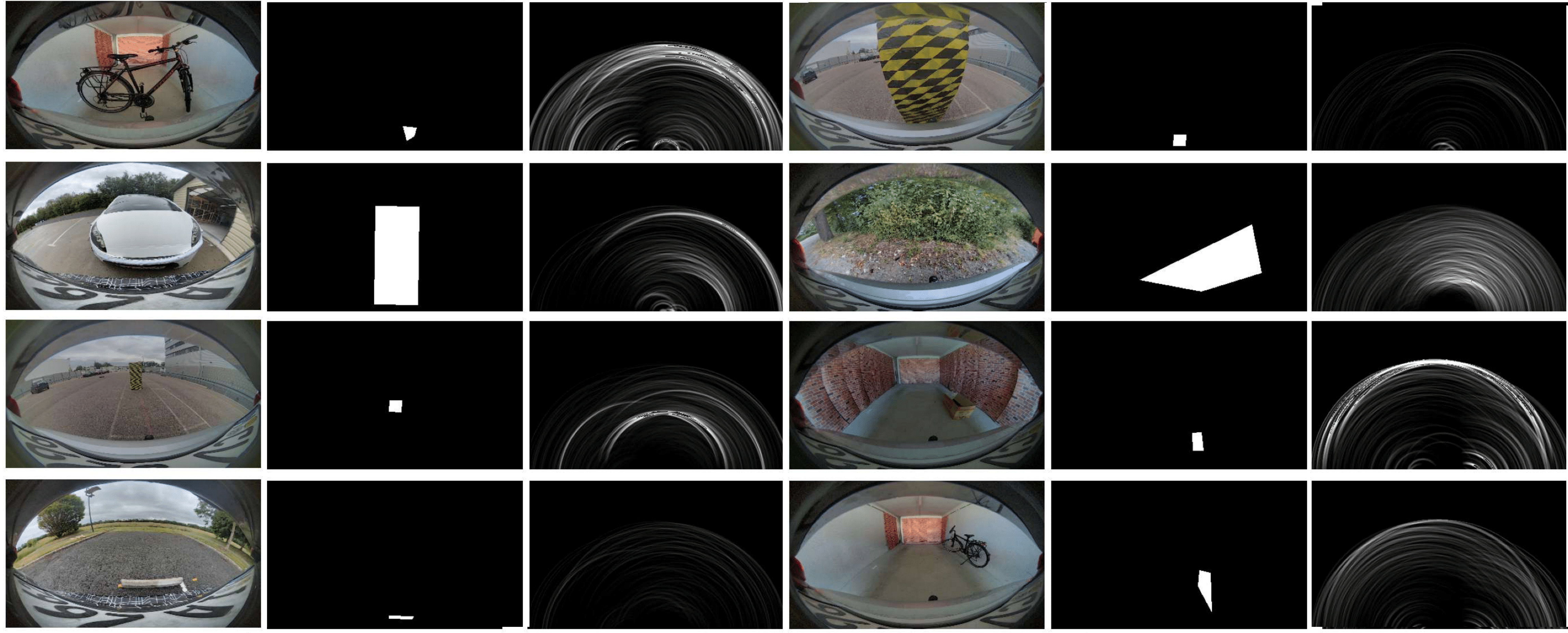}\\
    \caption{\textbf{Sample multimodal images and annotations from our custom dataset.} Fisheye images captured from the rear-view camera (first and fourth column); Corresponding ultrasonic BEV maps shown in the third and sixth columns; Segmentation masks of obstacles projected in bird's-eye-view perspective can be seen in the second and fifth columns.}

    \label{fig:obstacle_type}
\vspace{-4mm}
\end{figure}

%% file: include/architecture.tex
\section{Proposed Method} 

\begin{figure*}
    \captionsetup{singlelinecheck=false, font=small, belowskip=-6pt}
    \centering
    \includegraphics[width=\textwidth]{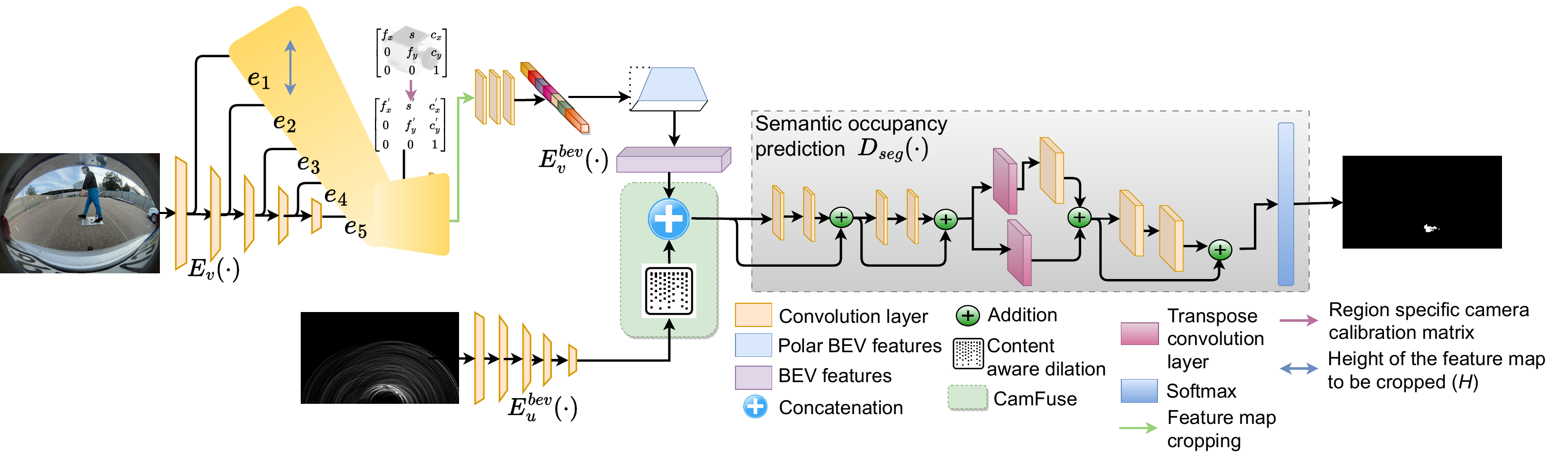}\\
    \caption{\textbf{Our proposed architecture for end-to-end multimodal obstacle perception in BEV.} The input fisheye RGB \& ultrasonic images are passed through two unimodal ResNext-50 \cite{xie2017aggregated} based feature encoders $E_v(\cdot)$ \& $E^{bev}_u(\cdot)$ respectively. A feature pyramid style network \cite{lin2017feature} is built on top of $E_v(\cdot)$ where the output features of each pyramid ($e_1, e_2,...$) are cropped to represent a specific region in the world with the help of area-specific estimated camera calibration matrix ($c_i'$). Further, the cropped features are flattened, converted to polar BEV features, and then to orthographic BEV features ($E^{bev}_v(\cdot)$). Ultrasonic-specific features from $E^{bev}_u(\cdot)$ are passed to the CamFuse module for content-aware dilation followed by multimodal fusion operation. Finally, multimodal features are fed to a two-stage semantic occupancy grid prediction $D_{seg}(\cdot)$ module followed by a $softmax$ layer.}
    \label{fig:proposed_method}
\vspace{-3mm}
\end{figure*}

The schematic representation of the proposed methodology is depicted in Figure \ref{fig:proposed_method}. To independently extract features from each modality, two separate unimodal encoders are employed. Utilizing the visible spectrum, the obtained unimodal features are transformed into BEV space. Subsequently, these features are fused with the features from the ultrasonic-specific encoder using the CamFuse module. Finally, the resultant multimodal bird's-eye-view features undergo processing through a two-stage deep semantic occupancy decoder. This iterative decoding refines the fused features, enhancing the precision of obstacle prediction for each grid. This proposal builds upon the foundational work conducted by Roddick \etal \cite{roddick2020predicting}.

\subsection{Unimodal encoders:} Obstacle types considered in this study are significantly diverse in terms of semantic representation in the visible spectrum. In addition, object occupancy in BEV for some of the obstacle types is not adequate to receive a strong response from ultrasonic sensors. Figure \ref{fig:obstacle_type} demonstrates different obstacle types in BEV and its corresponding ultrasonic responses. We use ResNeXt-50 \cite{xie2017aggregated} as one instance of the unimodal encoder ($E_v$) to exclusively extract the features from the visible spectrum in perspective view. Another such instance of the encoder ($E_u^{bev}$) is used to learn the ultrasonic-specific features in BEV.

Primarily, the feature distribution $\mathcal{F}$ of $E_v$ is in perspective view. As the next stage of this approach includes the transformation of these perspective features into BEV, we build a feature pyramid-style network \cite{lin2017feature}. Leveraging the available camera calibration parameters, each pyramid is tailored to cover a specific distance in the BEV, as elaborated in the subsequent section. One feature pyramid block is denoted by $e$ that is created from $E_v(\cdot)$. We have a series of such pyramid blocks 
$e= \{ {e_1}, {e_2}, \dots {e_N} \}$, $ {e_i} \subset {E_v}, {e_i} \in \mathbb{R}^{\mathcal{F}}$ where $N$ is the number of pyramids blocks that represent each multiscale feature. Each pyramid block ($e_i$) is derived by employing a set of residual blocks to extract and fine-tune the features in multiple stages. In addition, a feature block ($e_i$) which is in lower resolution is first upsampled and then concatenated with the feature maps of the immediately previous block ($e_{i-1}$). This helps to regularize the network to efficiently learn the contextual features at multiple spatial dimensions.

\subsection{Bird's-eye-view projection:}
In the BEV projection, ensuring the accurate representation of a particular grid region in the world coordinate space when reprojected back onto the feature map of its corresponding pyramid block is crucial. For this reason, the feature map from each feature pyramid block (\(e_i\)) undergoes a cropping process to determine the precise upper ($v_{max}$) and lower ($v_{min}$) bounds corresponding to the maximum bounds defined in actual world space. 
We use fisheye cameras that follow the Kannala-Brandt camera model \cite{kannala2006generic}. In contrast to characterizing radial distortion based on the distance from the image center (radius), the Kannala-Brandt model considers distortion as a function (\ref{eq:distortion_mtx}) of the incidence angle of the light passing through the lens. This approach is chosen because the distortion function is smoother when parameterized with respect to this angle ($\theta$), which makes it easier to model as a power-series in (\ref{eq:power_series}). 
\begin{equation}
    \pi(x,i) = \begin{bmatrix} f_xd(\theta) \frac{x}{r} \\ f_yd(\theta) \frac{y}{r} \end{bmatrix} + \begin{bmatrix} c_x \\ c_y \end{bmatrix}
    \label{eq:distortion_mtx}
\end{equation}
\begin{equation}
    r = \sqrt{x^2 + y^2}
    \label{eq:pyth_eq}
\end{equation}
\begin{equation}
    \theta = \text{atan2}(r, z)
\end{equation}                                  
\begin{equation}
    d(\theta) = \theta + k_1 \theta^3 + k_2 \theta^5 + k_3 \theta^7 + k_4 \theta^9
    \label{eq:power_series}
\end{equation}
where $f_x$ and $f_y$ are the focal lengths, r is the distance between the image point and the principal point. {$\theta$} is the angle between the principal axis and the incoming ray; $c_x$ and $c_y$ represent the principal points along the x and y axis; $k_i$ is the fisheye distortion parameters of $i^{th}$ order in the polynomial equation $d(\theta)$ and $z$ is the distance of the point from the camera.

To crop a region from the feature map, we determine distortion coefficients using equations (\ref{eq:pyth_eq} to \ref{eq:power_series}). Using equation (\ref{eq:distortion_mtx}) with corresponding focal length, distortion parameters, principal points, and world coordinate space heights and depths, we obtain a 2D image space coordinate ($u$, $v$) representing object boundaries in the world coordinate space. This transformation projects the minimum and maximum depths (\(z_{{min}}, z_{{max}}\)) and heights (\(y_{{min}}, y_{{max}}\)) for each grid onto 2D image space coordinates (common width $u$, maximum ($v_{max}$), minimum ($v_{min}$). This is then utilized for cropping the feature maps along the vertical axis. This vertical cropping is applied to each pyramid block $e= \{ {e_1}, {e_2}, \dots {e_N} \}$, where $N$ is considered as 5 in this study. The cropped feature map of each pyramid block covers a certain distance in the real world, which is as follows ${e_1}$: 3.2 to 6 meters, ${e_2}$: 1.6 to 3.2 meters, ${e_3}$: 0.8 to 1.6 meters, ${e_4}$: 0.4 to 0.8 meters and ${e_5}$: 0.2 to 0.4 meters respectively.

These feature maps, characterized by 256 channels, are then passed through a convolution block. This block, comprising a 2D convolution layer with an input of 256 channels and an output of 16 channels, is complemented by batch normalization and a subsequent ReLU activation block. The intricacy of this process lies in the reduction of dimensionality within the feature pyramid blocks, thereby enhancing their interpretability and utility in subsequent stages of the network. Furthermore, the output of this convolution block, constituting a 16-channel feature map, is subjected to further transformation through a dense 1D convolution layer. The input and output channel dimensions of this layer are contingent on the height, depth, and region coverage of the feature map, effectively mapping the feature representation from the image space to the Polar-BEV space. Finally, the feature maps from the Polar-BEV space are converted to an orthographic birds-eye-view space using a transformation, resembling the one in 
 \cite{roddick2020predicting}.

\subsection{Content-aware dilation and multimodal feature fusion (CaMFuse):}

CaMFuse utilizes modality-specific unimodal BEV features extracted from fisheye ($E_v^{bev}$) and ultrasonic ($E_u^{bev}$) space to synthesize modality-agnostic features. This can be achieved only when the domain gap can be minimized during the feature fusion. A primary challenge arises from the distinct representations of the environment by the considered sensors. The fisheye camera captures intricate semantic details in pixel format, incorporating geometric distortion, while the ultrasonic sensor perceives the surroundings through the reception of echoes generated by signal propagation. This discrepancy in sensing modalities adds complexity to achieving seamless fusion in representing the environment effectively. Moreover, potential sensor misalignment may occur, leading to the same object being positioned in two different grids within the two feature spaces. If this misalignment is not corrected, the resulting fused features could become ambiguous, compromising the accuracy of predictions. Addressing and rectifying such misalignments are crucial for ensuring the reliability and precision of the fused features in the final prediction process.

CaMFuse handles the issue highlighted above by adapting dilated convolution where the dilation operation is parameterized for the input data, as followed in \cite{yao2022adcnn}. Content-aware dilation is employed in the BEV feature space of ultrasonic data to account for its sparsity. It enables convolutional kernels to adapt their dilation values according to the distinct content present at the pixel level. We consider this operation to be performed for the features of ultrasonic data. Let us consider $E_u^{bev}$ constitutes of total $l$ number of convolution layers. The output, at $(i,j)$ location, denoted by $\mathcal{\lambda}$ using the dilated convolution kernel $\mathcal{K}$ for $l^{th}$ layer is defined as follows.
 \begin{equation} \label{eqn_entrphy}
	\lambda^l_{i,j} = \sum_{m=0}^{\mathcal{K}} \sum_{n=0}^{\mathcal{K}} w_{m,n} \times \mathcal{X}^{l-1}_{i+\delta^m,j+\delta^n}
\end{equation}
$\mathcal{X}^{l-1}$ is the input to the $(l-1)^{th}$ layer where $\mathcal{X}^{l-1} \in \mathbb{R}^{w_{l-1} \times h_{l-1}} : w^{l-1}, h^{l-1}$ are width and height corresponding to the input $\mathcal{X}^{l-1}$. In addition, $\delta$ and $w$ are the dilation value and weights of the learnable kernel $\mathcal{K}$.

The concept underlying adaptive dilated convolution involves estimating $\delta$ at the position $(i,j)$ using a function $\Delta_{i,j}$ that takes into account the input, represented as random variables $\mathcal{H(\cdot)}$, where $\delta_{i,j}$ is drawn from softmax sampling $\delta_{i,j} \sim softmax_{G}(\mathcal{H}_{i,j}(\cdot), \tau)$ \cite{JangGP17} with a smoothness parameter $\tau$ that controls the output being close to a true categorical distribution, inspired by \cite{hu2019learning}.
 \begin{equation} \label{eqn_entrphy}
	\delta_{i,j} = \Delta_{i,j}(\mathcal{H(\cdot)}) = \frac{e^{((\mathcal{H}_{i,j}(\cdot) + g_{i,j})/\tau)}}{\sum e^{((\mathcal{H}_{i,j}(\cdot) + g_{i,j})/\tau)}}
\end{equation}
Where $\mathcal{H}_{i,j}(\cdot)$ represents the set of underlying location specific hidden priors of $\mathcal{H}(\cdot)$; $g_{i,j} \overset{\text{i.i.d.}}{\sim} softmax_{G}(0,1) : g_{i,j} \in \mathbb{R}^D$ where $D$ indicates the number of valid options for dilation value. The hidden prior for $l^{th}$ layer is estimated by considering $\beta$ hierarchical layers followed by a recurrent aggregation method as defined below.
 \begin{equation} \label{eqn_entrphy}
	\mathcal{H}^l_{i,j} = f(w^l_{h} \cdot \mathcal{H}^{l-1}_{i,j} + U^l_{h} \cdot \beta^{l-1}_{i,j})
\end{equation}
Where, $w^l_h$ represents the weight matrix associated with the hidden prior in the previous layer, denoted as $\mathcal{H}^{l-1}_{i,j}$; $U^l_h$ corresponds to the weight matrix for the input in the $l^{th}$ layer; and $f(\cdot)$ denotes a non-linear activation function. The provided equation facilitates the aggregation of information from each layer as $l$ progresses deeper, fostering strong interdependence among the layers. However, as an extreme case of recurrent aggregation Yao \etal \cite{yao2022adcnn} applied Markov aggregation that set the kernel weights of previous hidden prior ($w^l_{h}$) as zero. This means similar to the Markov model \cite{gagniuc2017markov} only the last layer is enough to estimate the hidden prior of the content and learn the weights of the dilated convolution kernel.
 \begin{equation} \label{eqn_entrphy}
	\mathcal{H}^l_{i,j} = f( U^l_{h} \cdot \beta^{l-1}_{i,j})
\end{equation}
Content-aware dilated convolution is utilized on the BEV feature maps derived from the ultrasonic space ($E_u^{bev}$). Subsequently, these processed BEV feature maps are combined through concatenation with the feature maps originating from the fisheye stream ($E_v^{bev}$). This integration allows for a comprehensive fusion of information, leveraging content-aware dilated convolution to enhance the representation of the ultrasonic and fisheye features in the combined map.

\subsection{Semantic occupancy prediction:} We formulate a two-stage multi-scale semantic occupancy grid prediction decoder $\mathcal{D}_{Seg}(.)$ that takes the BEV feature maps generated by \textit{CaMFuse} as its input and produces predictions of occupancy in grid map \cite{elfes2013occupancy}.

We adopt the top-down approach designed to forecast the final probabilities of semantic occupancy in the grid from a set of BEV feature maps. The decoder is composed of two sequential residual blocks. The primary objective of the initial block is to avoid compromising spatial features within the same resolution. On another side, the second block operates on a higher dimension to exploit the strong context of automotive scenes and learn the prior geometry of diverse obstacle types. The initial block includes a set of successive convolution layers, succeeded by a concatenation operation. This sequence is iterated twice, maintaining the number of feature maps and resolution unchanged, as shown in Figure \ref{fig:proposed_method}. Notably, this block takes 16 feature maps as input and produces 32 feature maps as output. The next residual block commences with two parallel transpose convolution layers. One layer operates with a stride of 2 and kernel size of $4\times4$, while the other one employs a stride of 2 and kernel size of $2\times2$. However, both layers produce an output of 16 feature maps each. The output feature maps of the transpose layer with stride 1 are fed to a convolution layer with a stride of 1 and kernel size of $3\times3$ and produce the same number of features as input. Then the output of this layer and the output of the transpose layer with kernel size of $2\times2$ are concatenated and passed through two consecutive convolution layers with the same input and output channels with stride of 1 and kernel size of $3\times3$. The resulting feature maps are further concatenated with the feature maps obtained from the preceding concatenation operation. The feature maps undergo a concluding stage, traversing two more convolution layers {$\mathcal{C}onv_{1\times1}$}. Finally, we estimate the entropy ($\mathsf{E}$) between the distributions ground truth and prediction, as defined below.
\begin{equation} \label{entropy}
    \mathsf{E}[p(\cdot | x; \Omega)] \overset{\Delta}{=} -\sum_{i=1}^{m} p(y_i | x; \Omega)\log\left(\frac{\exp(\Phi_{\Omega}(x))}{\sum_{j=1}^{C} \exp(\Phi_{\Omega}(x))}\right)
\end{equation}
Where $\Omega$ denotes the model parameters. In addition, $\Phi_{\Omega}$ is a function, parameterized by $\Omega$, that takes $x$ as input, and ${y}_{i}$ represents the ground truth. This generates feature maps corresponding to the number of classes, which is 2 in our scenario, effectively discerning between obstacles and the background.

\definecolor{LightCyan}{rgb}{0.88,1,1}
\definecolor{LightGreen}{rgb}{0.35,0.8,0}
\definecolor{LightGreen2}{rgb}{0.5,0.9,0}
\definecolor{Gray}{gray}{0.85}

%% file: include/results.tex
\section{ Details of the experimental procedure}  \label{sec:performance}

\definecolor{LightCyan}{rgb}{0.88,1,1}
\definecolor{LightGreen}{rgb}{0.35,0.8,0}
\definecolor{LightGreen2}{rgb}{0.5,0.9,0}
\definecolor{Gray}{gray}{0.85}

\subsection{Evaluation metrics}
The task of obstacle perception involves binary segmentation, requiring the distinction between two classes: the background and the obstacles. We use a few standard metrics and one customized metric suitable for our application, they are discussed below.

\textbf{Recall} shows the system's ability to avoid critical misses, which is an essential measure in ADAS. \textbf{Dice} score offers an informative evaluation of object segmentation by considering object boundaries, handling class imbalances, promoting precise localization, and quantifying the degree of overlap. \textbf{Precision} score indicates where the system mistakenly identifies non-object regions as obstacles or other objects. So we also consider the precision score to determine the performance of our model. \textbf{Intersection over Union (IoU)} metric provides additional insights compared to other metrics like precision and recall by emphasizing spatial accuracy, handling false positives and false negatives, allowing for threshold flexibility, and directly assessing object completeness and shape as it'll have a direct impact on safe motion planning \cite{claussmann2019review}.

\textbf{Euclidean distance $(\mathcal{E})$}, also cited as ``E.D.", metric helps assess how closely the predicted obstacle and the ground truth obstacle align in terms of spatial location. The distance between the predicted obstacle and the actual obstacle is critical information to ensure that the system accurately perceives the location of obstacles.
\begin{equation}
    \mathcal{E} = \frac{1}{N} \sum_{i = 0}^{N} ((x_i^{t} - x_i^{p} )^2 + (y_i^{t} - y_i^{p} )^2)^{1/2}
\end{equation}
Where, $(x_i^{t}, y_i^{t})$ and $(x_i^{p}, y_i^{p})$ are the coordinates of the $i^{th}$ target and predicted obstacles among $N$ number of instances, ${N\in\mathbb{R}: N\geq0}$.

\textbf{Absolute distance error ($\mathcal{D}$)} metric, later referred to as ``Distance" which reflects how close or far the network perceives the target with respect to the vehicle as a reference. Understanding this relative distance helps in making real-time decisions about object avoidance, braking, or steering. This metric is defined in the equation below.
\begin{equation}
    \mathcal{D} = \frac{1}{N} \sum_{i = 0}^{N} |v_i - \hat{v}_i|
\end{equation}
Where, $v_i$ and $\hat{v}_i$ are the $Y$ positions of the $i^{th}$ target and predicted obstacle among $N$ such instances ${N\in\mathbb{R}: N\geq0}$.

\textbf{Normalized distance ($\mathcal{N}_D$)}, later shortened as ``N. Distance" or ``N. D.", metric as determined by the Euclidean distance between the predicted obstacle location and the actual target obstacle divided by the corresponding Euclidean distance between the target obstacle and the ego vehicle. The rationale behind formulating this metric comes from the fact that while other key performance indicators (KPIs) measure the model's accuracy in obstacle perception, it is imperative, given the unique context of this study, to assess the model's performance with the ego vehicle as a reference point. The main intuition is that the errors in obstacle localization nearer to the ego vehicle carry more impact than errors associated with obstacles situated at a greater distance. This metric is defined as below.
\begin{equation}
    \mathcal{N}_D = \frac{1}{N} \sum_{i = 0}^{N} \frac{((x_i^{t} - x_i^{p} )^2 + (y_i^{t} - y_i^{p} )^2)^{1/2}}{((x_i^{t} - x^{e} )^2 + (y_i^{t} - y^{e} )^2)^{1/2}}
\end{equation}
Where, $x_i^{t}$, $x_i^{p}$ are the $X$ positions and $y_i^{t}$, $y_i^{p}$ are the $Y$ positions of the $i^{th}$ target and predicted obstacles among $N$ number of instances, ${N\in\mathbb{R}: N\geq0}$. In addition, $(x^{e}, y^{e})$ is the coordinate of the ego vehicle.

In summary, recalls, dice score, precision, and IoU are interpreted as better when they are higher, while absolute distance error, normalized distance, and Euclidean distance are better when they are lower.

\subsection{Implementation details}
We present an architectural framework that applies to both multimodal and unimodal solutions, at a high level characterized by two unimodal encoders and a common decoder block. 
For the unimodal training with visible data, CamFuse is omitted, as it solely deals with RGB inputs. Given the absence of a specialized dataset for ultrasonic-based unimodal training, we encountered a challenge as BEV maps in the visible spectrum, which typically serve as annotations for the multimodal network, could not be directly employed due to the lack of semantic information within ultrasonic sensor data. To address this limitation, we estimated the center coordinates of obstacles from the existing semantic annotations and utilized them as the target output annotations for ultrasonic-based unimodal training.

\begin{table*}[!h]
\centering
\captionsetup{font=small}
\caption{Quantitative comparison of \textbf{unimodal vs. multimodal models in two different environments}.}
\scalebox{0.9}{
\begin{tabular}{ll|c|c|c|c|c|c|c}
\toprule
\textbf{Environment}                        & \textbf{Modality} & \textbf{Recall $\uparrow$} & \textbf{Dice Score $\uparrow$} & \textbf{Precision $\uparrow$} & \textbf{IoU $\uparrow$} & \textbf{Distance ($\mathcal{D}$) $\downarrow$} & \textbf{N. Distance ($\mathcal{N}_D$) $\downarrow$} & \textbf{E. D. $(\mathcal{E})$ $\downarrow$} \\ \midrule
\multicolumn{1}{c}{\multirow{3}{*}{Indoor}} & Visible           &         0.5078        &       0.4739              &         0.9899           &          0.4469                &                   0.3486                  &            0.9762                 &                  2.0131                  \\
\multicolumn{1}{c}{}                        & Ultrasonic        &         -        &            -         &         -           &            -              &     1.6578                     &              0.1410               &       2.0668                             \\
\multicolumn{1}{c}{}                        & Multimodal        &         0.9430        &         0.7649            &        0.9430            &         0.6518                 &           0.1147                          &              0.0314               &                0.1312                    \\ \midrule
\multicolumn{1}{c}{\multirow{3}{*}{Outdoor}} & Visible           &       0.5997          &         0.5948            &        0.7195            &        0.5122                  &      0.3352                   &     0.4255                        &          1.0984                          \\
                                            & Ultrasonic        &       -          &         -            &           -         &             -             &           0.8717             &        0.0776                     &                    1.0790                \\
                                            & Multimodal        &       0.9089          &         0.8069            &        0.9310            &           0.7141               &            0.0424                    &      0.0300                       &               0.1369                     \\ \bottomrule
\end{tabular}
}
\label{tab:environment}
\end{table*}

We used PyTorch to develop all the models and conducted training for 100 epochs. Adam optimizer \cite{kingma2014adam} with a base learning rate of $1 \times 10^{-3}$ was employed. We applied categorical cross entropy as the loss function for both multimodal and RGB-specific unimodal training, while mean squared error (MSE) was used for the ultrasonic-specific unimodal model. All training sessions were conducted on an NVIDIA GeForce RTX 2080 Ti GPU, with a batch size of 8 for all experiments.

\begin{table}[!h]
\centering
\captionsetup{font=small}
\caption{\textbf{Ablation study of the proposed architecture on our dataset}. ResNet-50 \cite{he2016deep} and ResNeXt-50 \cite{xie2017aggregated}, recognized as standard backbones, serve as unimodal feature extractors for both visible and ultrasonic data. Three distinct fusion techniques, including the proposed method, are systematically assessed. The experimentation encompasses four diverse loss functions for each fusion approach: Dice, Mean Square Error (MSE), Binary Cross Entropy (BCE), and Categorical Cross Entropy (CCE).
}
\scalebox{0.8}{
\begin{tabular}{l|c|c|c|c|c|c|c}
\toprule

\multicolumn{1}{l|}{\textbf{Encoders}} & \multicolumn{1}{l|}{\textbf{Fusion}}          & \textbf{Loss} & \textbf{Recall $\uparrow$}  & \textbf{Precision $\uparrow$} & \textbf{IoU $\uparrow$}  & \textbf{N. D.  $\downarrow$} & \textbf{E. D. $\downarrow$} \\ \midrule
\multirow{12}{*}{\rotatebox{90}{ResNet-50 \cite{he2016deep}}}             & \multirow{4}{*}{$f_1$}                
                                        & Dice          &  0.5741	&	0.6836	&	0.4798	&	1.312	&	1.8905                       \\
                                                        &                                               
                                        & MSE           &  0.7345	&	0.7951	&	0.6078	&	0.892	&	1.2081                     \\
                                        &                                               
                                        & BCE           &  0.7702	&	0.8197	&	0.6118	&	0.612	&	1.1762                      \\
                                        &                                               
                                        & CCE           &  0.7987	&	0.8325	&	0.6338	&	0.382	&	0.9856                       \\ \cmidrule(r){2-8}
                                        & \multirow{4}{*}{$f_2$}              
                                        & Dice          &  0.6387	&	0.6163	&	0.4602	&	0.9144	&	1.3178                      \\
                                        &                                               
                                        & MSE           &  0.8068	&	0.8568	&	0.4725	&	0.8754	&	0.917                      \\
                                        &                                               
                                        & BCE           &  0.8349	&	0.8764	&	0.5001	&	0.7182	&	0.8378                       \\
                                        &                                               
                                        & CCE           &  0.8584	&	0.9447	&	0.5162	&	0.6124	&	0.6878                       \\ \cmidrule(r){2-8}
                                        & \multirow{4}{*}{$f_3$}              
                                        & Dice          &  0.5972	&	0.7221	&	0.5179	&	0.831	&	1.1451                      \\
                                        &                                               
                                        & MSE           &  0.7653	&	0.8185	&	0.5897	&	0.8706	&	0.9276                      \\
                                        &                                               
                                        & BCE           &  0.8152	&	0.9166	&	0.5995	&	0.3452	&	0.6243                      \\
                                        &                                               
                                        & CCE           &  0.8824	&	0.9102	&	0.6024	&	0.2365	&	0.5357                      \\ \midrule
\multirow{12}{*}{\rotatebox{90}{ResNeXt-50 \cite{xie2017aggregated}}}             & \multirow{4}{*}{$f_1$}                
                                        & Dice          &  0.7248	&	0.7747	&	0.283	&	0.1872	&	0.9636                      \\
                                        &                                               
                                        & MSE           &  0.7544	&	0.8642	&	0.388	&	0.1577	&	0.8747                      \\
                                        &                                               
                                        & BCE           &  0.7846	&	0.8693	&	0.433	&	0.1373	&	0.8323                      \\
                                        &                                               
                                        & CCE           &  0.8248	&	0.8844	&	0.478	&	0.1422	&	1.0025                      \\ \cmidrule(r){2-8}
                                        & \multirow{4}{*}{$f_2$}              
                                        & Dice          &  0.7403	&	0.7512	&	0.3214	&	0.259	&	0.6729                      \\
                                        &                                               
                                        & MSE           &  0.8505	&	0.7816	&	0.5117	&	0.164	&	0.5329                      \\
                                        &                                               
                                        & BCE           &  0.8657	&	0.8869	&	0.5413	&	0.119	&	0.4571                      \\
                                        &                                               
                                        & CCE           &  0.8755	&	0.8864	&	0.5468	&	0.094	&	0.4325                      \\ \cmidrule(r){2-8}
                                        & \multirow{4}{*}{$f_3$}              
                                        & Dice          &  0.7017	&	0.6215	&	0.4683	&	0.3047	&	0.5458                      \\
                                        &                                               
                                        & MSE           &  0.7616	&	0.7117	&	0.5181	&	0.2295	&	0.4253                      \\
                                        &                                               
                                        & BCE           &  0.8513	&	0.8016	&	0.5485	&	0.1696	&	0.3502                      \\
                                        &                                               
                                        & CCE           &  \textbf{0.9115}	&	\textbf{0.9361}	&	\textbf{0.6537}	&	\textbf{0.0792}	&	\textbf{0.2151}                      \\ \bottomrule                        
\end{tabular}
}
\label{tab:ablation}
\vspace{-5mm}
\end{table}

\begin{table*}[!h]
\centering
\captionsetup{font=small}
\caption{\textbf{Obstacle type specific comparative study} of accurate perception using both - unimodal and multimodal models.}
\scalebox{0.83}{
\begin{tabular}{ll|c|c|c|c|c|c|c}
\toprule
\textbf{Obstacle Type}                        & \textbf{Modality} & \textbf{Recall $\uparrow$} & \textbf{Dice Score $\uparrow$} & \textbf{Precision $\uparrow$} & \textbf{IoU $\uparrow$} & \textbf{Distance ($\mathcal{D}$) $\downarrow$} & \textbf{N. Distance ($\mathcal{N}_D$) $\downarrow$} & \textbf{E. D. $(\mathcal{E})$ $\downarrow$} \\ \midrule
\multicolumn{1}{c}{\multirow{3}{*}{Dummy Pedestrian}} & Visible           &      0.4659           &       0.5404              &  0.4659        & 0.4046        &   0.0274     &   0.0710      &   0.3350          \\
\multicolumn{1}{c}{}                        & Ultrasonic        &       -  &     -          &  -           &    -           &    0.2806           &   0.0286              &  0.3953                           \\
\multicolumn{1}{c}{}                        & Multimodal        &   0.9642     &    0.8096       &   0.9642      &      0.7240     &   0.0302    &     0.0196       &    0.1058                     \\ \midrule
\multicolumn{1}{c}{\multirow{3}{*}{Carton}} & Visible           &       0.5620   &    0.5340     &   0.9927   &    0.5108   &     0.8216      &    0.8701         &   1.9150                          \\
                                            & Ultrasonic        &     -        &      -       &    -         &     -         &   1.6763      &    0.1476        &      2.0613                   \\
                                            & Multimodal        &    0.7893     &    0.7202     &     0.8268      &     0.6278  &      0.0962   &     0.0322    &     0.1466                       \\ \midrule
\multicolumn{1}{c}{\multirow{3}{*}{Bicycle}} & Visible           &    0.000     &   0.000     &    0.0000      &   0.000       &      0.6597     &     2.000       &    4.0794                         \\
                                            & Ultrasonic        &      -        &    -       &      -       &      -        &     0.8705     &    0.0838        &    1.0419                         \\
                                            & Multimodal        &     0.9048    &   0.6339     &    0.9048       &    0.4769   &     0.1980     &    0.0556       &     0.2221              \\ \midrule
\multicolumn{1}{c}{\multirow{3}{*}{Round Pillar}} & Visible     &     0.000     &    0.000     &   0.000        &    0.000       &     1.0300      &   2.000        &    4.5985                  \\
                                            & Ultrasonic        &     -        &      -       &      -        &     -        &    0.0707     &     0.0063        &     0.0834                     \\
                                            & Multimodal        &    0.8554     &   0.6459     &   0.9342     &     0.5092     &      0.0286  &    0.0174         &    0.0788                         \\ \midrule
\multicolumn{1}{c}{\multirow{3}{*}{Bottle Case}} & Visible      &    0.9861       &     0.9166        &    0.9861     &    0.8614    &    0.0230    &     0.0041     &       0.0259         \\
                                            & Ultrasonic        &      -     &     -     &   -       &   -        &     1.9536       &   0.1626     &    2.5013           \\
                                            & Multimodal        &    0.9931    &   0.9099    &    0.9931    &    0.8366   &    0.0319  &    0.0061  &    0.0405           \\ \midrule
\multicolumn{1}{c}{\multirow{3}{*}{Square Pillar}} & Visible    &   1.0000     &    0.9558     &   1.0000    &     0.9180    &     0.0180     &    0.0048   &    0.0260                   \\
                                            & Ultrasonic        &      -    &   -          &    -         &      -        &    0.5052         &    0.0438    &   0.6179             \\
                                            & Multimodal        &   1.0000     &    0.9457     &    1.0000    &   0.8903   &    0.0170        &    0.0047    &   0.0256             \\ \midrule
\multicolumn{1}{c}{\multirow{3}{*}{Car}} & Visible        &   1.0000    &    0.7589       &   0.9859    &    0.6189    &    0.0436     &     0.2001     &    0.8656       \\
                                            & Ultrasonic        &     -       &   -        &     -       &    -     &    0.3440      &    0.0296   &   0.4396            \\
                                            & Multimodal        &   0.9859    &   0.8599   &    0.9859  &    0.7702  &   0.0771    &   0.0744      &   0.3574           \\ \midrule
\multicolumn{1}{c}{\multirow{3}{*}{Vegetation}} & Visible       &   0.1333       &    0.3244     &    0.1333   &   0.1974    &   0.1116    &   1.1640   &   2.8718          \\
                                            & Ultrasonic        &    -      &     -      &     -    &    -     &   0.9016    &   0.0819    &   1.1365           \\
                                            & Multimodal        &   1.0000   &    0.7615 &   1.0000  &   0.6346  &   0.1433  &   0.1839    &   0.6238          \\ \midrule
\multicolumn{1}{c}{\multirow{3}{*}{All}} & Visible              &  0.5184     &    0.5038    &   0.5705    &   0.4389     &    0.3419     &     0.7893   &   1.8397             \\
                                            & Ultrasonic        &  -          &    -         &     -       &     -        &    0.8253     &     0.0730   &   1.0346             \\
                                            & Multimodal        &  0.9366     &    0.7858    &   0.9511    &   0.6837     &    0.0778     &     0.0492   &   0.2001             \\ \bottomrule

\end{tabular}
}
\label{tab:obstacle_type}
\vspace{-3mm}
\end{table*}

\subsection{Ablation study}
Although our proposed methodology is modular and supports both unimodal and multimodal networks, we focus the ablation study on the multimodal model. We design an extensive range of experiments on our dataset to evaluate the impact of the proposed fusion technique of the multimodal model with other standard backbones and find out the best suitable loss function to train the same.

\textbf{Network configuration:}
Three different fusion strategies are compared, marked as $f_1$ $\rightarrow$ simple concatenation of feature maps from two different modalities, $f_2$ $\rightarrow$ feature fusion as described in \cite{dasgupta2022spatio} followed by concatenation and $f_3$ $\rightarrow$ proposed CamFuse method. 
These fusion strategies are systematically explored in tandem with four diverse and suitable loss functions: Dice, Mean Square Error (MSE), Binary Cross Entropy (BCE), and Categorical Cross Entropy (CCE). Each fusion-loss combination is meticulously tested and evaluated. To ensure a comprehensive analysis, these combinations are further applied using ResNet-50 \cite{he2016deep} and ResNeXt-50 \cite{xie2017aggregated} as unimodal feature extractors, adding an additional layer of complexity and nuance to the experimentation.
Table \ref{tab:ablation} shows this ablation study where ResNeXt-50 as a backbone with the proposed fusion strategy using categorical cross-entropy loss turns out to be the best network setup of the proposed method.

\textbf{Data Augmentation:}
Data Augmentation is an essential aspect of any deep learning-based algorithms to efficiently regularize the model by bringing diversity to the training distribution. In this work, in the visible spectrum, we apply simple augmentation techniques such as enhancing brightness, contrast, saturation from a range of 0.8 to 1.2, and hue [-0.1, 0.1]. The impact of this set of augmentations can be observed in Table \ref{tab:obstacle_type} corresponding obstacle type ``All" while comparing with the best KPI candidate from Table \ref{tab:ablation}.

\begin{table*}[!h]
\centering
\captionsetup{font=small}
\caption{Comparison of \textbf{unimodal vs. multimodal models on identification and localization of obstacles} with respect to its various positions.}
\scalebox{0.85}{
\begin{tabular}{ll|c|c|c|c|c|c|c}
\toprule
\textbf{Obstacle Position}                        & \textbf{Modality} & \textbf{Recall $\uparrow$} & \textbf{Dice Score $\uparrow$} & \textbf{Precision $\uparrow$} & \textbf{IoU $\uparrow$} & \textbf{Distance ($\mathcal{D}$) $\downarrow$} & \textbf{N. Distance ($\mathcal{N}_D$) $\downarrow$} & \textbf{E. D. $(\mathcal{E})$ $\downarrow$} \\ \midrule
\multicolumn{1}{c}{\multirow{3}{*}{Eccentric}} & Visible        &   0.3867   &  0.4874 &    0.4966  &   0.4096  &    0.2464  &   0.4867     &     1.2233        \\
\multicolumn{1}{c}{}                        & Ultrasonic        &       -    &   -     &    -    &     -     &   0.8375      &   0.0698     &     0.9580        \\
\multicolumn{1}{c}{}                        & Multimodal        &   0.9440   &  0.8155 &    0.9620  &   0.7193  &   0.0281   &   0.0153     &    0.0796         \\ \midrule
\multicolumn{1}{c}{\multirow{3}{*}{Central}} & Visible          &   0.7250   &  0.6265 &    0.9091  &   0.5468  &   0.1638   &   0.5209     &      1.2108       \\
                                            & Ultrasonic        &   -        &    -    &    -     &      -    &    0.9191    &   0.0711     &     1.0339        \\
                                            & Multimodal        &   0.9600   & 0.8166  &   0.9600  &  0.7094   &  0.0588    &    0.0354    &  0.1513           \\ \midrule
\multicolumn{1}{c}{\multirow{3}{*}{Corner}} & Visible           &   1.0000   & 0.7984  &  1.0000 &  0.6684    &  0.0848     &  0.3176       &    1.2776         \\
                                            & Ultrasonic        &       -    &   -     &    -    &    -      &   0.8347      &   0.0739     &    1.0972         \\
                                            & Multimodal        &   1.0000   & 0.8608  &  1.0000 &  0.7636   &  0.0989      &    0.0904     &    0.4350         \\ \bottomrule

\end{tabular}
}
\label{tab:obstacle_position}
\end{table*}

\begin{table*}[!h]
\centering
\captionsetup{font=small}
\caption{\textbf{Obstacle distance categorized comparison study of unimodal and multimodal proposals} for BEV-based obstacle perception.}
\scalebox{0.85}{
\begin{tabular}{ll|c|c|c|c|c|c}
\toprule
\textbf{Obstacle Distance [meter]}                        & \textbf{Modality} & \textbf{Recall $\uparrow$} & \textbf{Dice Score $\uparrow$} & \textbf{Precision $\uparrow$} & \textbf{IoU $\uparrow$} & \textbf{N. Distance ($\mathcal{N}_D$) $\downarrow$} & \textbf{E. D. $(\mathcal{E})$ $\downarrow$} \\ \midrule
\multicolumn{1}{c}{\multirow{3}{*}{0 - 1.45}} & Visible         &   0.8204   &   0.7557  &   0.8535 &  0.6742   &   0.1287   &      0.5238    \\
\multicolumn{1}{c}{}                        & Ultrasonic        &   -        &    -      &   -      &   -       &  0.0949    &    1.5409      \\
\multicolumn{1}{c}{}                        & Multimodal        &   0.9397   &   0.7732  &   0.9397 &  0.6846   &  0.0382    &    0.2177      \\ \midrule
\multicolumn{1}{c}{\multirow{3}{*}{1.45 - 2.9}} & Visible       &   0.6375   &   0.5766  &   0.8430 &  0.4997   &   0.5765   &    1.5500      \\
                                            & Ultrasonic        &   -        &   -       &   -      &    -      &   0.0807   &    1.1926      \\
                                            & Multimodal        &   0.7484   &   0.6574  &   0.7987  &  0.5632  &   0.0299   &    0.1492     \\ \midrule
\multicolumn{1}{c}{\multirow{3}{*}{2.9 - 4.35}} & Visible       &   0.5796   &   0.5439  &   0.8239 &  0.4790   &   0.7173   &    1.5909       \\
                                            & Ultrasonic        &    -       &   -       &   -       &   -      &   0.0694   &    0.9764       \\
                                            & Multimodal        &   0.9207   &   0.7578  &   0.9248  &  0.6529  &   0.0488   &    0.1882       \\ \midrule
\multicolumn{1}{c}{\multirow{3}{*}{4.35 - 5.8}} & Visible       &   0.4501   &   0.4761  &   0.7066  &  0.4231  &   0.7587   &    1.6208       \\
                                            & Ultrasonic        &    -       &    -      &    -      &    -     &   0.0622   &    0.8147       \\
                                            & Multimodal        &   0.9624   &   0.8274  &   0.9746  &  0.7397  &   0.0194   &    0.0793       \\ \bottomrule

\end{tabular}
}
\label{tab:distance}
\vspace{-5mm}
\end{table*}

\subsection{Quantitative study}
In this section, we present a comprehensive quantitative analysis to assess the effectiveness of the proposed approach. To the best of our knowledge, our work represents the first effort in creating a multisensor solution that integrates a fisheye camera and an ultrasonic sensor for rear-view obstacle perception. As no prior art exists for direct comparisons, we rely on our custom dataset to present and discuss the obtained results, given the unavailability of a publicly suitable dataset tailored to our specific use case.

\textbf{Environment} plays a crucial role in any computer vision algorithms as it comes with diverse lighting conditions between different scenarios as shown in Figure \ref{fig:different_obstacles}. In Table \ref{tab:environment}, we provide a comparative analysis of our proposed approach in both indoor and outdoor scenarios. Our assessment includes the performance of both unimodal and multimodal models. Our results suggest the superior performance of the model trained with multimodal data across all evaluated metrics. However, recall, dice score, precision, and IoU are not applicable for the model trained with only ultrasonic input as the task was defined as a regression problem.

\textbf{Obstacle type} covers a wide range of categories in our dataset, as shown in Table \ref{tab:obstacle_type}, to account for the varied obstacles frequently documented in different reports. The heterogeneous geometric and semantic characteristics of distinct obstacles underscore the importance of a thorough evaluation of each object category. Such an approach provides valuable insights into the model's behavior and highlights instances of failure. An overview of the results presented in the table demonstrates that, across all obstacle types, the multimodal model outperforms unimodal alternatives. This superior performance is not confined to mere obstacle presence recognition; it also extends to precise localization, as evidenced by metrics such as normalized and Euclidean distance. It is important to highlight that for scenes featuring obstacles like bicycles and round pillars, the visible model encountered challenges in segmenting anything as an obstacle. This limitation is evident from the precision and IoU metrics. However, despite this, we have presented the distance-based metrics. The reason behind this choice is that we cannot consider the distance-based metrics to be 0, unlike other metrics, as a zero distance would imply complete overlap between the obstacle segmentation output and the ground truth, potentially leading to an incorrect interpretation. Therefore, we calculate the distance-based metric as the distance between the ground truth point and the farthest point possible within the camera range. This approach provides a more meaningful assessment of the model's performance in scenarios where obstacle segmentation is particularly challenging.

\begin{table*}[!h]
\centering
\captionsetup{font=small}
\caption{\textbf{Quantitative comparison of unimodal and multimodal proposals} on our custom dataset to validate the robustness of each model against ego motion.}
\scalebox{0.84}{
\begin{tabular}{ll|c|c|c|c|c|c|c}
\toprule
\textbf{Ego Speed [km/h]}                        & \textbf{Modality} & \textbf{Recall $\uparrow$} & \textbf{Dice Score $\uparrow$} & \textbf{Precision $\uparrow$} & \textbf{IoU $\uparrow$} & \textbf{Distance ($\mathcal{D}$) $\downarrow$} & \textbf{N. Distance ($\mathcal{N}_D$) $\downarrow$} & \textbf{E. D. $(\mathcal{E})$ $\downarrow$} \\ \midrule
\multicolumn{1}{c}{\multirow{3}{*}{0.5 - 3}} & Visible           &   0.5567  &   0.5439   &    0.8126    &   0.4898       &    0.3844     &     0.6931       &   1.5752                 \\
\multicolumn{1}{c}{}                        & Ultrasonic        &       -    &     -     &     -       &     -    &     1.3040     &    0.1125    &        1.6128                   \\
\multicolumn{1}{c}{}                        & Multimodal        &    0.9137  &  0.7890   &    0.9305   &   0.6904   & 0.0723      &     0.0314      &      0.1389                       \\ \midrule
\multicolumn{1}{c}{\multirow{3}{*}{3 - 5}} & Visible           &      0.5591 &    0.5710  &   0.6852   &  0.4921   &     0.3680   &    0.4669    &         1.1973                    \\
                                            & Ultrasonic        &       -    &     -     &     -       &    -     &     0.9768    &    0.0860   &     1.1969                        \\
                                            & Multimodal        &   0.9007  &  0.8025   &   0.9250   &   0.7106   &   0.0421     &    0.0311   &       0.1415           \\ \midrule
\multicolumn{1}{c}{\multirow{3}{*}{5 - 8}} & Visible           &     0.6507 &   0.6225  &    0.6507  &      0.4922  &   0.0345 &    0.0541    &        0.2605                     \\
                                            & Ultrasonic        &    -      &    -  &      -          &     -      &   0.2641    &   0.0284   &        0.3917             \\
                                            & Multimodal        &  0.9658   &  0.8161   &   0.9658    &    0.7081  &   0.0327   &    0.0208    &        0.1082                  \\ \bottomrule

\end{tabular}
}
\label{tab:speed}
\vspace{-2mm}
\end{table*}

\begin{figure}
    \captionsetup{singlelinecheck=false, font=small}
    \centering
    \includegraphics[width=1.14in, height=0.9in]{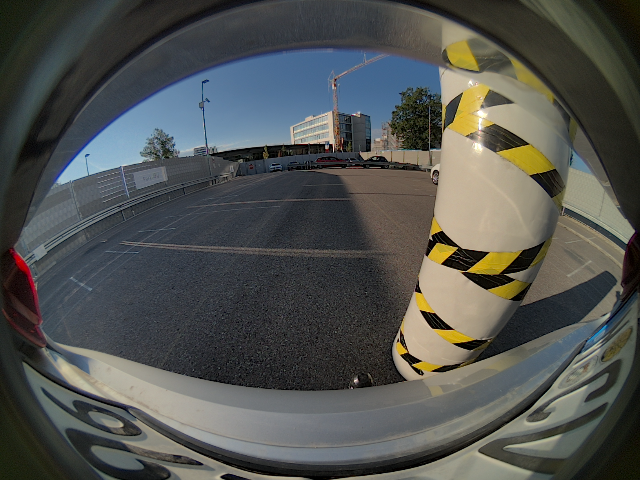}
    \hspace{-0.078in}
    \includegraphics[width=1.14in, height=0.9in]{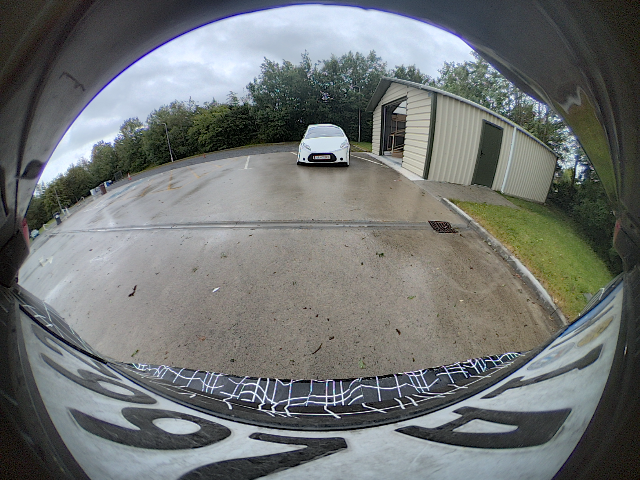}
    \hspace{-0.078in}
    \includegraphics[width=1.14in, height=0.9in]{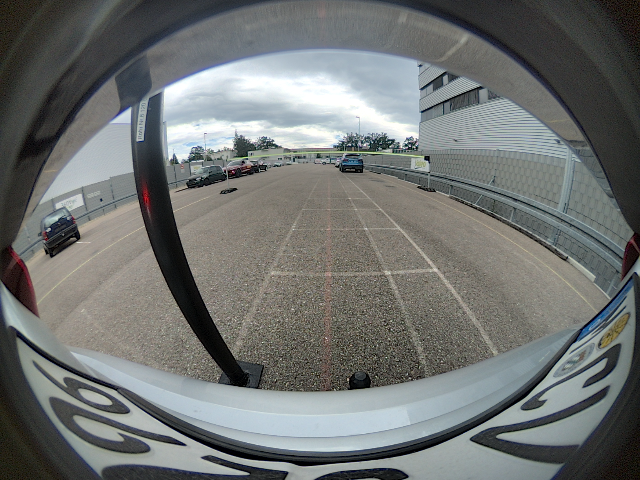}\\
    \vspace{1mm}
    \includegraphics[width=1.14in, height=0.9in]{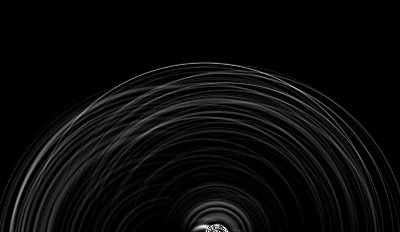}
    \hspace{-0.078in}
    \includegraphics[width=1.14in, height=0.9in]{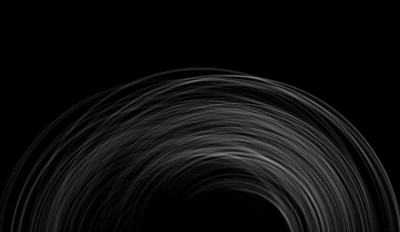}
    \hspace{-0.078in}
    \includegraphics[width=1.14in, height=0.9in]{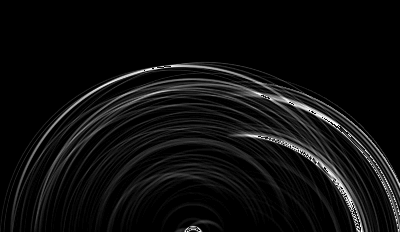}\\
    \caption{Samples exhibiting \textbf{different obstacle positions} - corner (left), central (middle), eccentric (right). Top: fisheye images; Bottom: ultrasonic BEV maps;
    }
    
    \vspace{-10mm}

    \label{fig:obstacle_position}
\end{figure}

\textbf{Obstacle position} is another aspect to validate the generalizability of the model because the semantic representation in the fisheye camera and response of the ultrasonic sensor differ based on the position of an object. In our dataset, we organize the position of obstacles into three distinct categories - corner (both extreme ends), central (exactly at the middle), and eccentric (anywhere else other than corner and central) respectively as illustrated in Figure \ref{fig:obstacle_position}. Evaluation specific to each category of obstacle position is presented in Table \ref{tab:obstacle_position} where the multimodal model turns out to be superior compared to the unimodal proposals and more accurate in localizing obstacles as indicated by Euclidean distance.

\textbf{Obstacle distance}, plays a critical role in assessing the model's precision in obstacle perception. Table \ref{tab:distance} shows a comprehensive set of metrics categorized into four different obstacle distances from the ego vehicle. Our findings show the superior performance of the multimodal model in comparison to the unimodal models across all measured parameters.

\textbf{Ego speed} makes a direct impact on the perception stack for all sensors due to the influence of the ego vehicle being in motion, which also causes blurring on the visible spectrum. To check how our multimodal model performs when the ego vehicle is in motion, we report the same set of metrics in three different ego speed categories, detailed in Table \ref{tab:speed}. In summary, the multimodal model exhibits (a) its superiority among all proposals, (b) equal effectiveness across all ego vehicle speeds, and (c) demonstrates its robustness against the motion while being extremely precise in localizing the obstacles.

\subsection{Qualitative study}
\begin{figure*}
    \captionsetup{singlelinecheck=false, font=small}
    \centering
    
    \raisebox{0.35in}{\rotatebox[origin=t]{90}{\scriptsize RGB input}}\includegraphics[width=1.14in, height=0.9in]{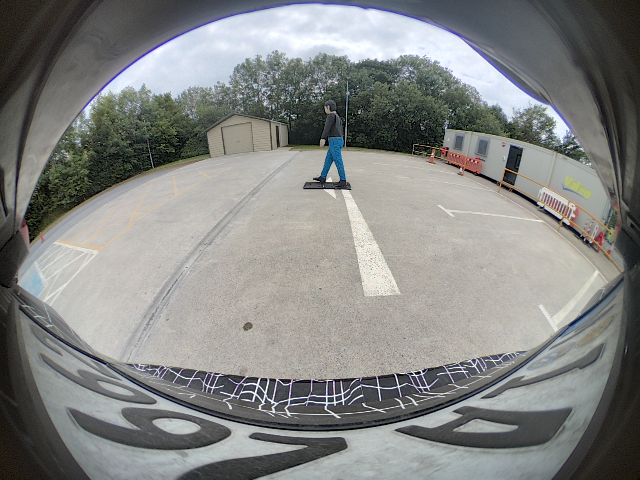}
    \hspace{-0.078in}
    \includegraphics[width=1.14in, height=0.9in]{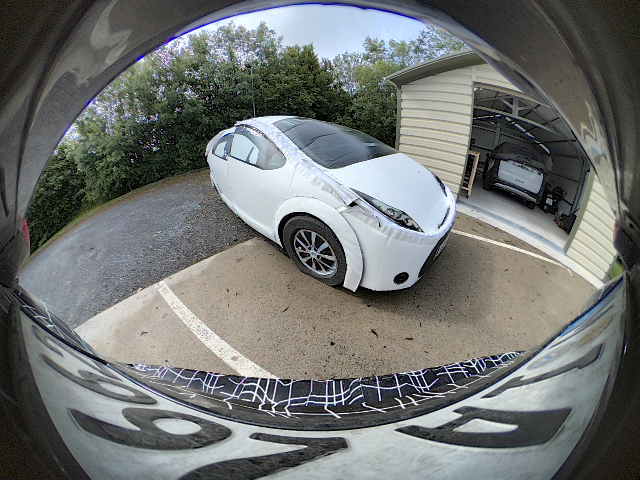}
    \hspace{-0.078in}
    \includegraphics[width=1.14in, height=0.9in]{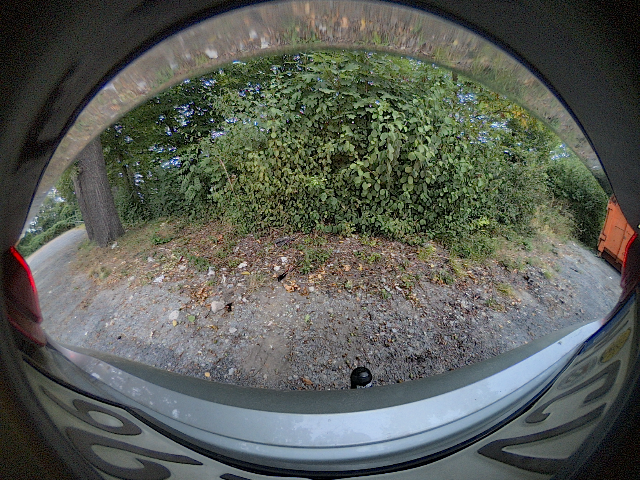}
    \hspace{-0.078in}
    \includegraphics[width=1.14in, height=0.9in]{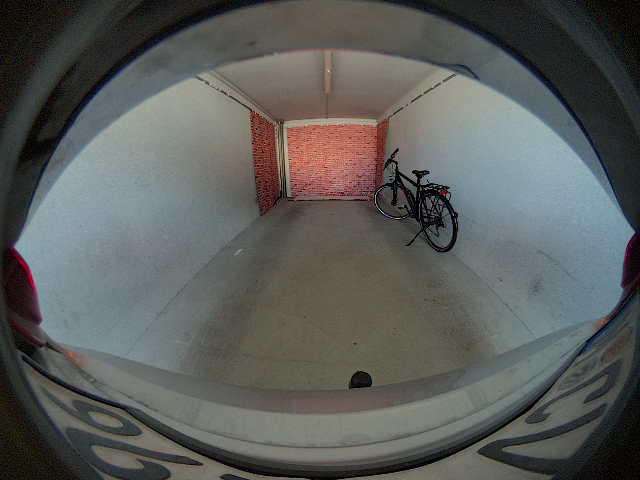}
    \hspace{-0.078in}
    \includegraphics[width=1.14in, height=0.9in]{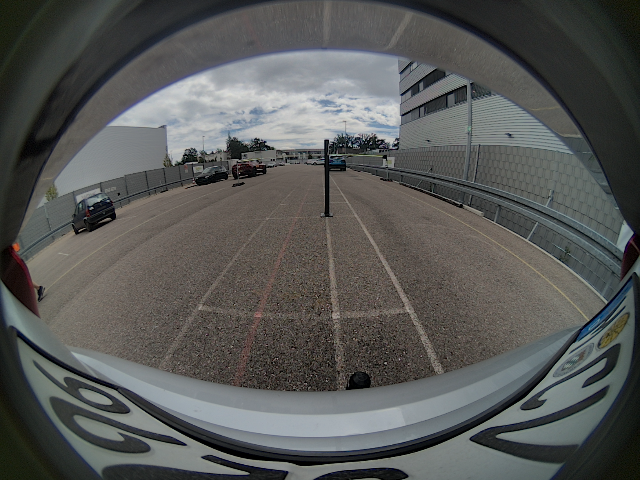}
    \hspace{-0.078in}
    \includegraphics[width=1.14in, height=0.9in]{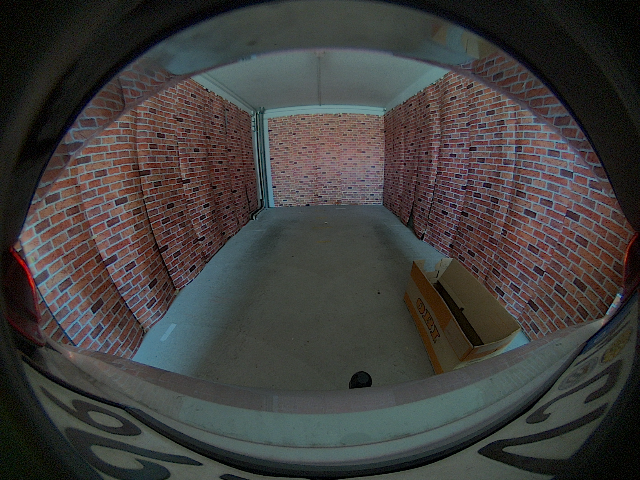}\\
    \vspace{1mm}
    \raisebox{0.35in}{\rotatebox[origin=t]{90}{\scriptsize Ultrasonic Input}}\includegraphics[width=1.14in, height=0.9in]{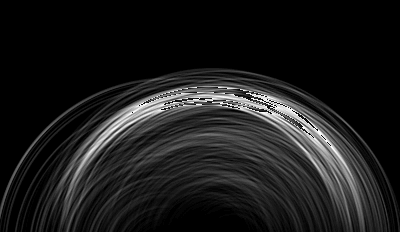}
    \hspace{-0.078in}
    \includegraphics[width=1.14in, height=0.9in]{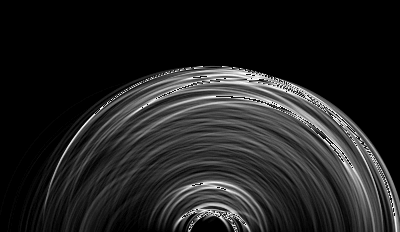}
    \hspace{-0.078in}
    \includegraphics[width=1.14in, height=0.9in]{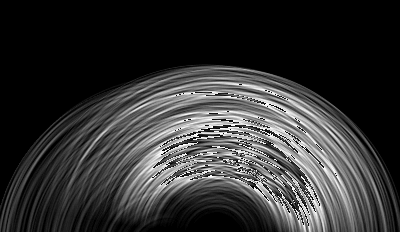}
    \hspace{-0.078in}
    \includegraphics[width=1.14in, height=0.9in]{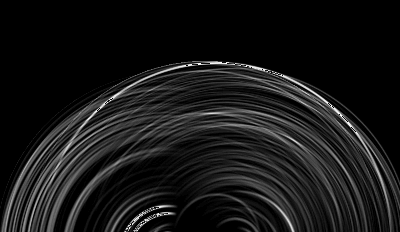}
    \hspace{-0.078in}
    \includegraphics[width=1.14in, height=0.9in]{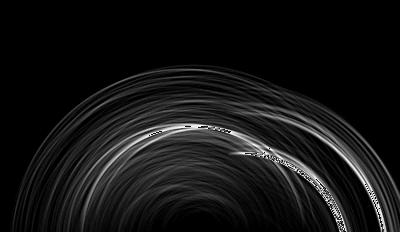}
    \hspace{-0.078in}
    \includegraphics[width=1.14in, height=0.9in]{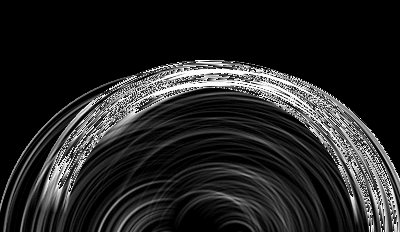}\\
    \vspace{1mm}
    \raisebox{0.38in}{\rotatebox[origin=t]{90}{\scriptsize Unimodal output \tiny (RGB)}}\includegraphics[width=1.14in, height=0.9in]{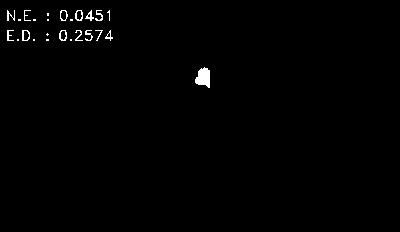}
    \hspace{-0.078in}
    \includegraphics[width=1.14in, height=0.9in]{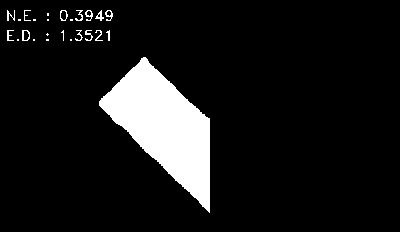}
    \hspace{-0.078in}
    \includegraphics[width=1.14in, height=0.9in]{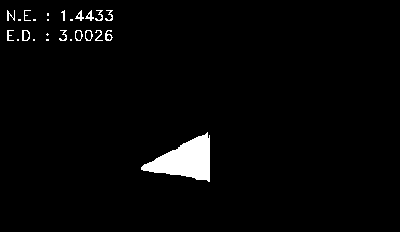}
    \hspace{-0.078in}
    \includegraphics[width=1.14in, height=0.9in]{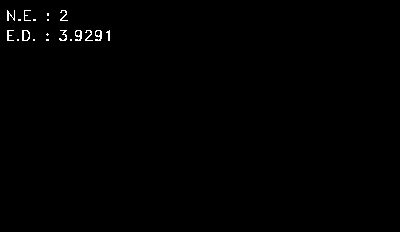}
    \hspace{-0.078in}
    \includegraphics[width=1.14in, height=0.9in]{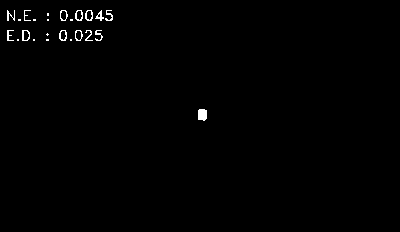}
    \hspace{-0.078in}
    \includegraphics[width=1.14in, height=0.9in]{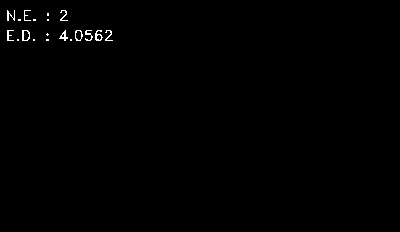}\\
    \vspace{1mm}
    \raisebox{0.38in}{\rotatebox[origin=t]{90}{\scriptsize Unimodal output \tiny (ULS)}}\includegraphics[width=1.14in, height=0.9in]{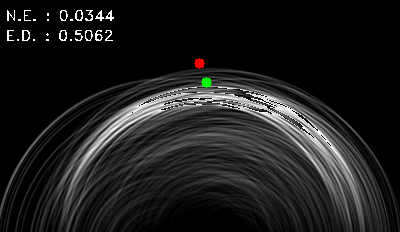}
    \hspace{-0.078in}
    \includegraphics[width=1.14in, height=0.9in]{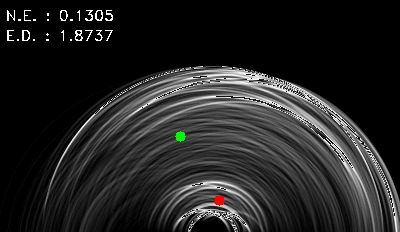}
    \hspace{-0.078in}
    \includegraphics[width=1.14in, height=0.9in]{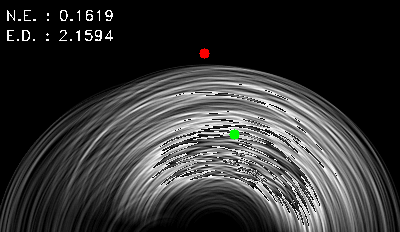}
    \hspace{-0.078in}
    \includegraphics[width=1.14in, height=0.9in]{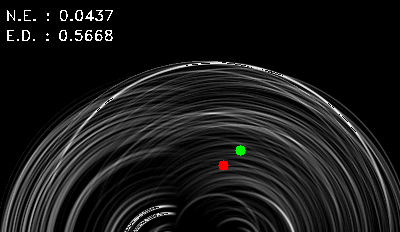}
    \hspace{-0.078in}
    \includegraphics[width=1.14in, height=0.9in]{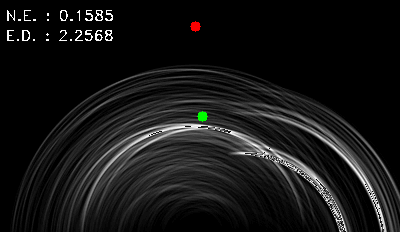}
    \hspace{-0.078in}
    \includegraphics[width=1.14in, height=0.9in]{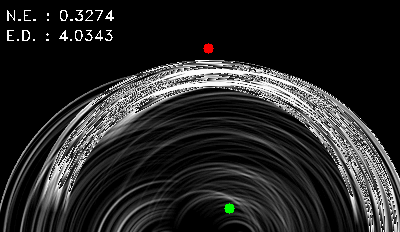}\\
    \vspace{1mm}
    \raisebox{0.38in}{\rotatebox[origin=t]{90}{\scriptsize Multimodal output}}\includegraphics[width=1.14in, height=0.9in]{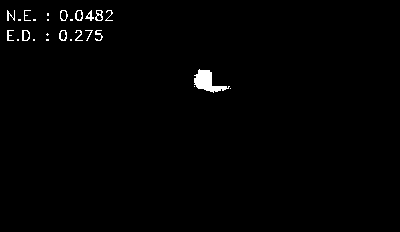}
    \hspace{-0.078in}
    \includegraphics[width=1.14in, height=0.9in]{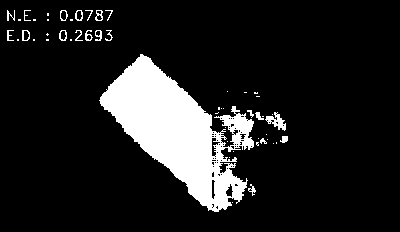}
    \hspace{-0.078in}
    \includegraphics[width=1.14in, height=0.9in]{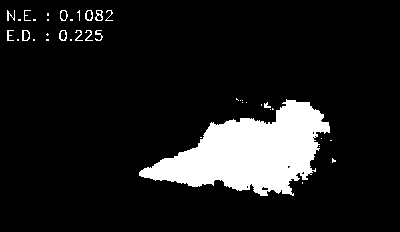}
    \hspace{-0.078in}
    \includegraphics[width=1.14in, height=0.9in]{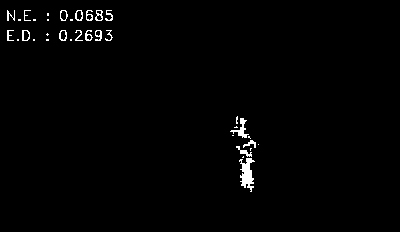}
    \hspace{-0.078in}
    \includegraphics[width=1.14in, height=0.9in]{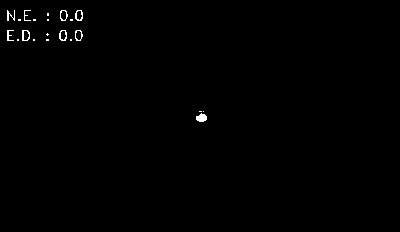}
    \hspace{-0.078in}
    \includegraphics[width=1.14in, height=0.9in]{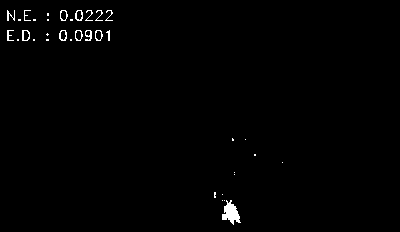}\\
    \vspace{1mm}
    \raisebox{0.38in}{\rotatebox[origin=t]{90}{\scriptsize BEV ground truth}}\includegraphics[width=1.14in, height=0.9in]{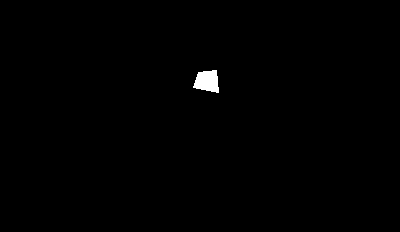}
    \hspace{-0.078in}
    \includegraphics[width=1.14in, height=0.9in]{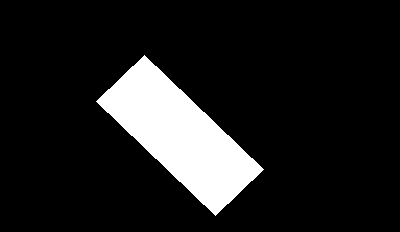}
    \hspace{-0.078in}
    \includegraphics[width=1.14in, height=0.9in]{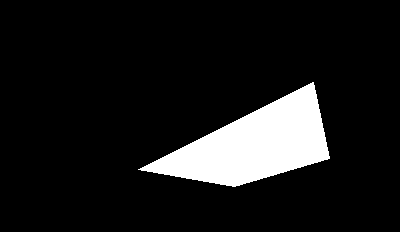}
    \hspace{-0.078in}
    \includegraphics[width=1.14in, height=0.9in]{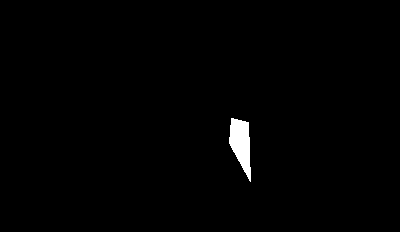}
    \hspace{-0.078in}
    \includegraphics[width=1.14in, height=0.9in]{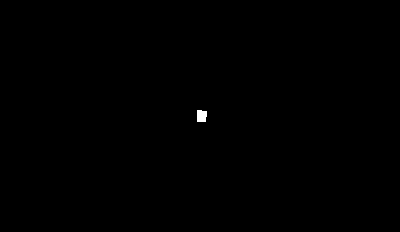}
    \hspace{-0.078in}
    \includegraphics[width=1.14in, height=0.9in]{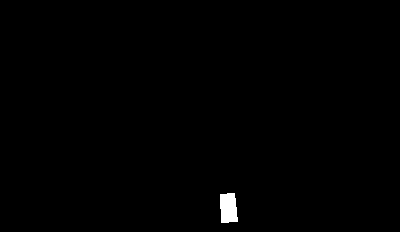}\\
    \vspace{1mm}
    \raisebox{0.38in}{\rotatebox[origin=t]{90}{\scriptsize Multimodal image output}}\includegraphics[width=1.14in, height=0.9in]{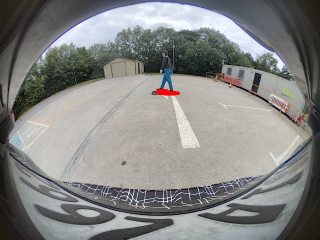}
    \hspace{-0.078in}
    \includegraphics[width=1.14in, height=0.9in]{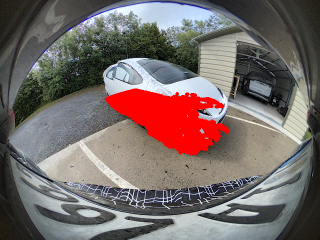}
    \hspace{-0.078in}
    \includegraphics[width=1.14in, height=0.9in]{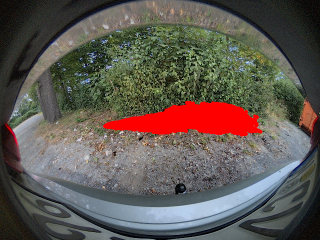}
    \hspace{-0.078in}
    \includegraphics[width=1.14in, height=0.9in]{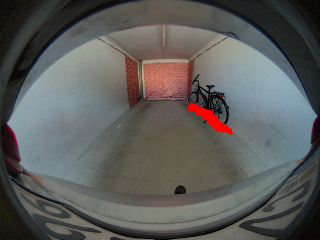}
    \hspace{-0.078in}
    \includegraphics[width=1.14in, height=0.9in]{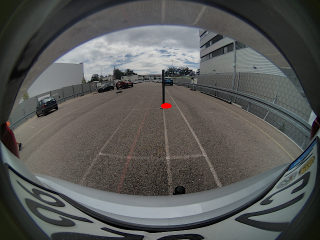}
    \hspace{-0.078in}
    \includegraphics[width=1.14in, height=0.9in]{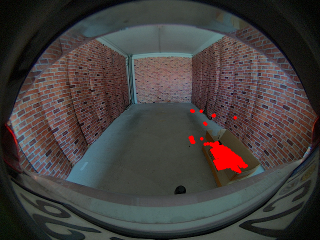}\\
    \vspace{1mm}
    \raisebox{0.38in}{\rotatebox[origin=t]{90}{\scriptsize Reprojected  GT}}\includegraphics[width=1.14in, height=0.9in]{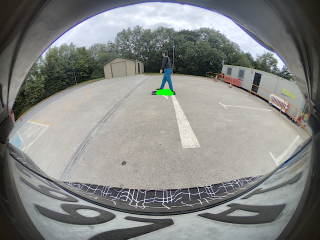}
    \hspace{-0.078in}
    \includegraphics[width=1.14in, height=0.9in]{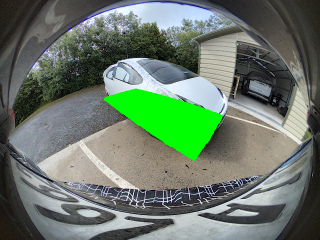}
    \hspace{-0.078in}
    \includegraphics[width=1.14in, height=0.9in]{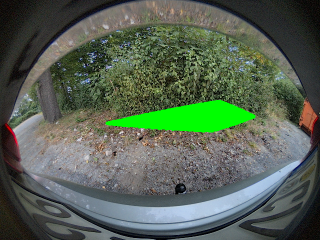}
    \hspace{-0.078in}
    \includegraphics[width=1.14in, height=0.9in]{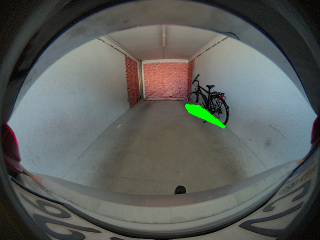}
    \hspace{-0.078in}
    \includegraphics[width=1.14in, height=0.9in]{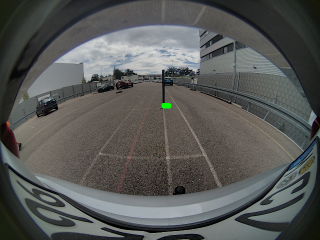}
    \hspace{-0.078in}
    \includegraphics[width=1.14in, height=0.9in]{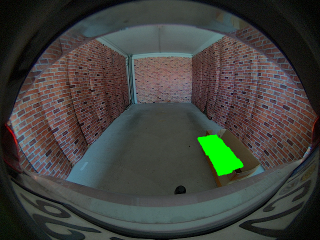}\\
    \caption{\textbf{Qualitative results of the proposed method}. \textit{First row} $\rightarrow$ input fisheye images; \textit{Second row} $\rightarrow$  input ultrasonic BEV maps; \textit{Third row} $\rightarrow$  BEV output from RGB only model; \textit{Fourth row} $\rightarrow$  BEV output from ultrasonic only model.  \textit{Fifth row} $\rightarrow$  BEV output from multimodal model; \textit{Sixth row} $\rightarrow$  BEV ground truth. Normalized error (N.E.) and Euclidean distance (E.D.) are highlighted for each unimodal and multimodal model at rows three, four, and five respectively. \textit{Seventh row} $\rightarrow$ Multimodal predictions from fifth row reprojected back to input fisheye image space; \textit{Eighth row} $\rightarrow$ BEV ground truth corresponding to sixth row reprojected back to input fisheye image space; \textcolor{red}{RED} and \textcolor{green}{GREEN} colors indicate prediction and ground truth;
    }
    \vspace{-5mm}

    \label{fig:all_results}
\end{figure*}

Figure \ref{fig:all_results} shows qualitative results of the proposed method along with models trained with unimodal inputs. 
There are cases where unimodal models based on RGB input completely fail to detect the obstacles, which can be seen in the third row of the fourth and sixth columns of Figure \ref{fig:all_results}. It is worth noting that in situations where the unimodal model succeeds in detecting the obstacle, it can inadvertently contribute to potentially hazardous ego-motion planning due to inadequacies in the segmentation output coverage. Conversely, another unimodal model trained using only ultrasonic BEV maps can show reasonable performance for localization (fourth row of first and fourth column), opposite behavior can be observed in other columns of fourth row. The performance of our proposed multimodal model across a wide range of scenarios, obstacle types, and distances consistently excels in detecting and accurately localizing obstacles. It surpasses the performance of unimodal models (rows $\rightarrow$ third and fourth) in terms of normalized error and Euclidean distance, providing a robust solution for obstacle perception on BEV. The multimodal predictions, along with their corresponding ground truths in BEV (fifth and sixth rows), are reprojected back to the input fisheye image space, allowing further visualization in the seventh and eighth rows. More qualitative results are provided at \url{https://youtu.be/JmSLBBL9Ruo}.

\begin{figure}
    \captionsetup{singlelinecheck=false, font=small}
    \centering
    \includegraphics[width=0.4\textwidth]{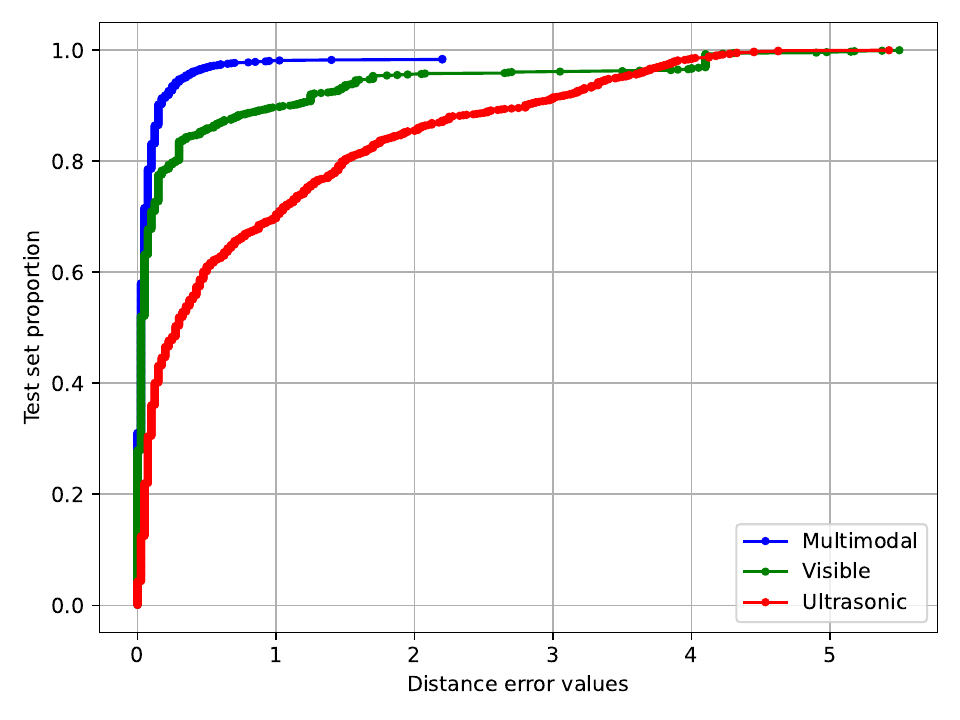}\\
    \caption{\textbf{Comparative analysis of models trained with multimodal, ultrasonic and visible data}, evaluated for absolute distance error metric on our test dataset.}
    \label{fig:distant_metric}
    \vspace{-10mm}
\end{figure}

\subsection{Unimodal vs. multimodal perception}
Multimodal perception, which combines information from different sensor modalities, consistently outperforms unimodal models across all tested scenarios and conditions. In Figure \ref{fig:distant_metric}, the cumulative absolute distance error plot illustrates that 90\% of the test samples exhibit nearly zero meters of absolute distance error. However, when considering models trained solely on visible sensor data, it displays an absolute distance error ranging between 1-2 meters and 3-4 meters for the unimodal model trained with only ultrasonic data. This validation confirms that models tailored to specific sensors lack the necessary generalization to consistently deliver high performance across various scenarios, especially when compared with a multimodal model. The benefits of multimodal perception are twofold: it enhances the ability of the system to detect and localize obstacles, and it significantly contributes to the robustness of the model in diverse environmental and motion conditions.

%% file: include/conclusions.tex
\section{Conclusion}  \label{sec:conc}
In this paper, we introduce a novel end-to-end deep learning architecture for multimodal obstacle perception in bird's-eye-view combining fisheye camera and ultrasonic sensor. We project fisheye camera images in BEV and then perform fusion with ultrasonic sensors using content-aware dilation and multimodal feature fusion module to reduce the domain gap between two sensors. We provide clear steps to create a similar multisensor dataset, data capture strategy, and sensor-specific preprocessing details. We also propose absolute distance error and normalized distance - two custom metrics to accurately evaluate the performance of obstacle perception in BEV. Our thorough experimental study of our internal dataset shows the significant superiority of the multimodal model over the unimodal proposals. In future work, we plan to bring other automotive sensors and fuse them with fisheye cameras and ultrasonic sensors to accomplish other related tasks for a complete surround-view system.

%% file: include/bio.tex

\begin{IEEEbiography}
[{\includegraphics[width=1in,height=1.5in,clip,keepaspectratio]{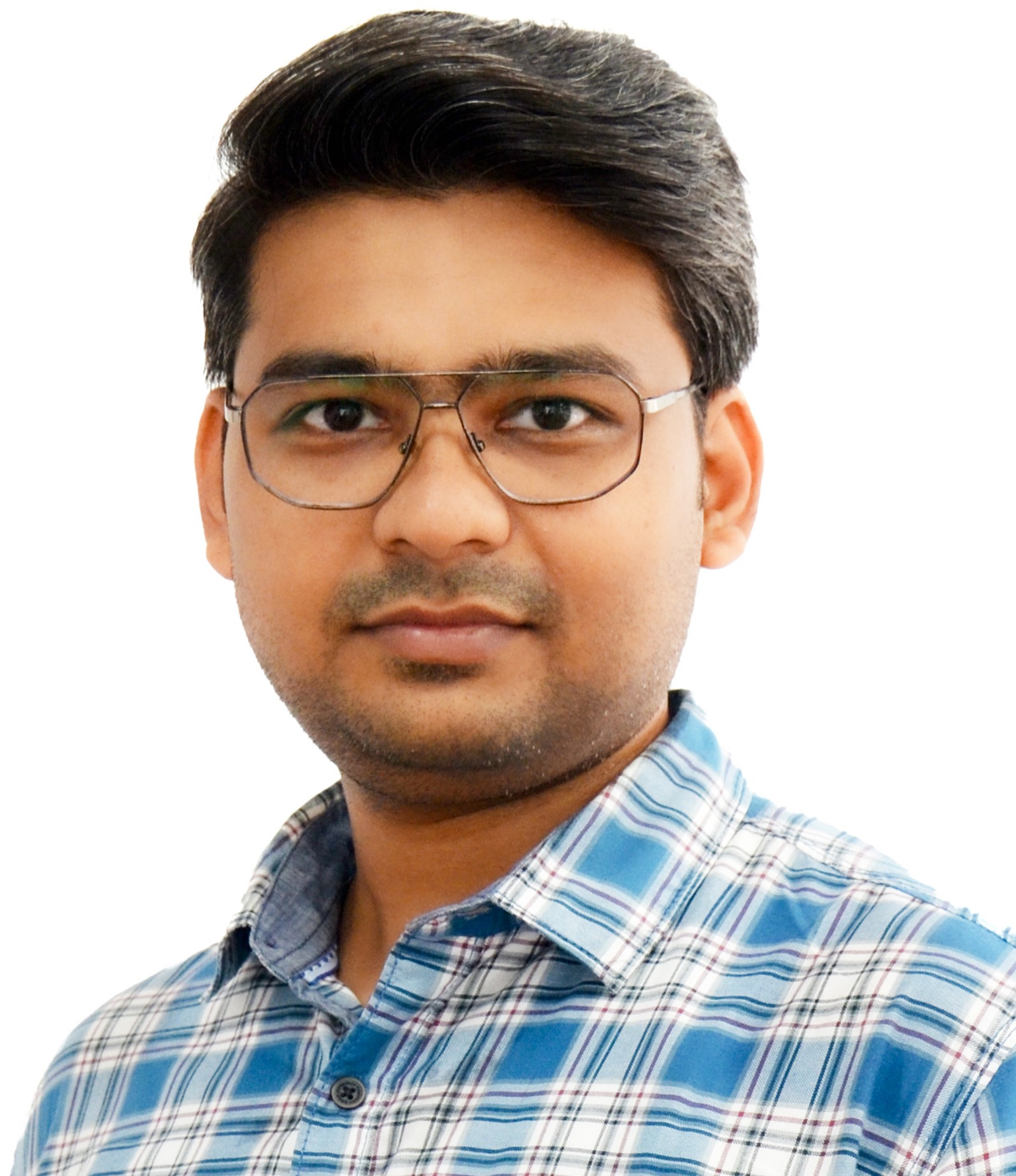}}]{Arindam Das} is currently pursuing the Ph.D. degree with the Department of Electronic and Computer Engineering, University of Limerick, Ireland. He is also an AI Software Architect with the Department of Driving Software and Systems (DSW), Valeo, India, where he holds the position of an Expert in AI. He is responsible to design AI algorithms to support various features for autonomous driving systems. He has 10 years of industry experience in computer vision, deep learning, and document analysis. He has authored 17 peer-reviewed publications and 52 patents. His current research interests include weakly-supervised learning, domain adaptation, image restoration, and multimodal learning.
\end{IEEEbiography}
\vskip -1\baselineskip plus -1fil

\begin{IEEEbiography}
[{\includegraphics[width=1in,height=1.5in,clip,keepaspectratio]{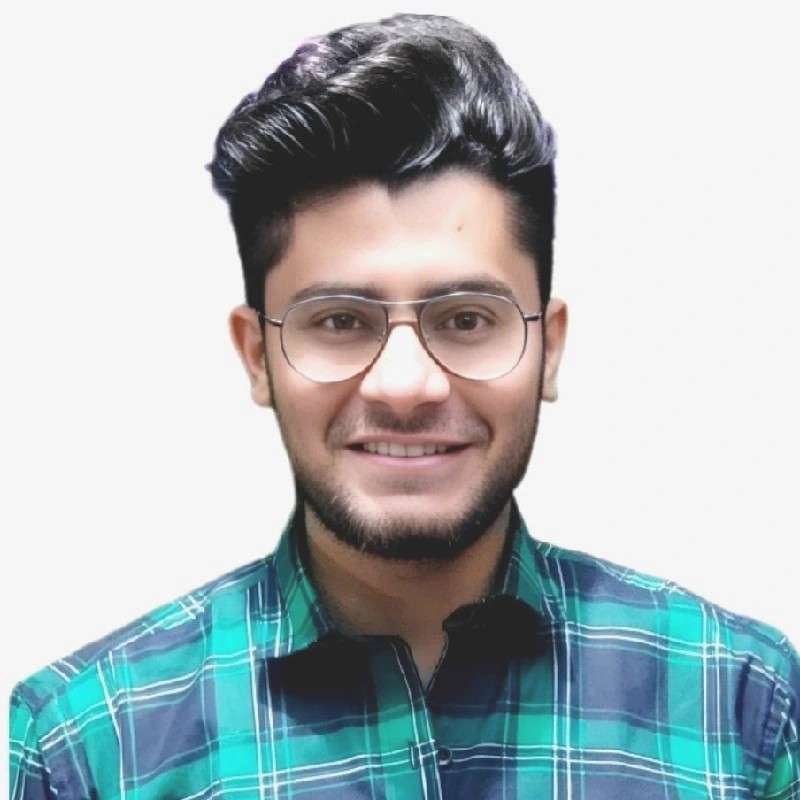}}]{Sudarshan Paul} is a Deep Learning research engineer in the Department of Driving Software and Systems (DSW), Valeo, India. He holds an Engineering degree in Electronics and Communication and has prior experience as Machine Learning Engineer at TCS with Smart Mobility team, where he specialized in various ADAS applications. He has also contributed to research activities in gaming controls using Computer Vision at Samsung R\&D Bangalore and behavioral modeling with Pattern Recognition at Tata Research Development and Design Centre, Pune. Currently, his research focus is on Multimodal Learning, Uncertainty Estimation, BEV Segmentation and detection for ADAS applications.
\end{IEEEbiography}
\vskip -1\baselineskip plus -1fil

\begin{IEEEbiography}
[{\includegraphics[width=1in,height=1.25in,clip,keepaspectratio]{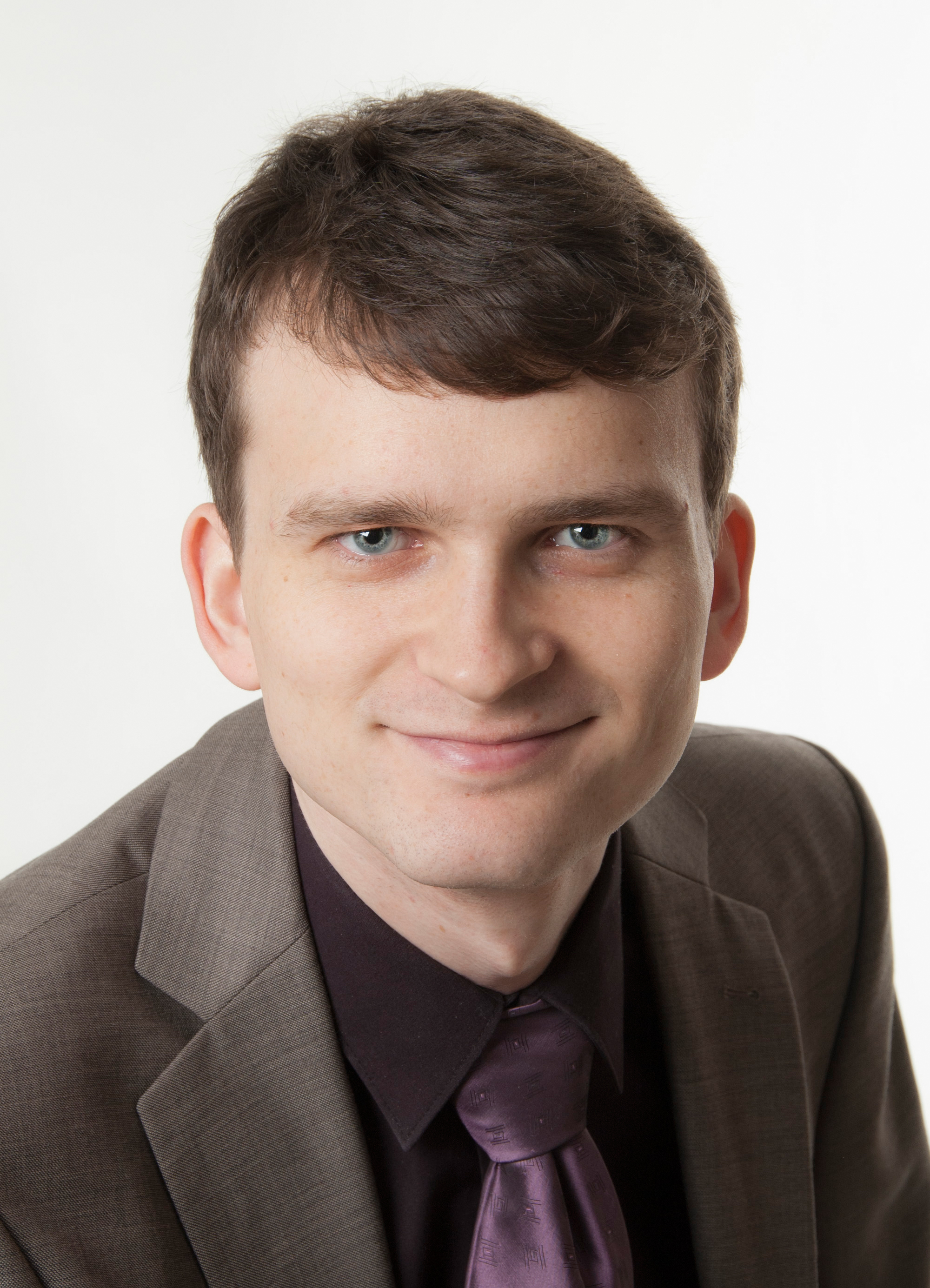}}]{Niko Scholz} is a Software Engineer at the Valeo Comfort and Driving Assistance site in Kronach, Germany. After finishing his Master of Science in Physics from University of Würzburg in 2019, he joined Valeo and worked on classical tracking algorithms as well as machine learning applications for ultrasonic data analysis. Later work also featured the application of Deep Neural Networks to Radar data. His current interests lie in the fusion of multiple sensing modes like radar, ultrasonic sensors and cameras in classical mapping and tracking applications as well as in Neural Network architectures to fuse them in feature space. 
\end{IEEEbiography} 
\vskip 1\baselineskip plus -1fil

\begin{IEEEbiography}[{\includegraphics[width=1in,height=1.25in,clip,keepaspectratio]{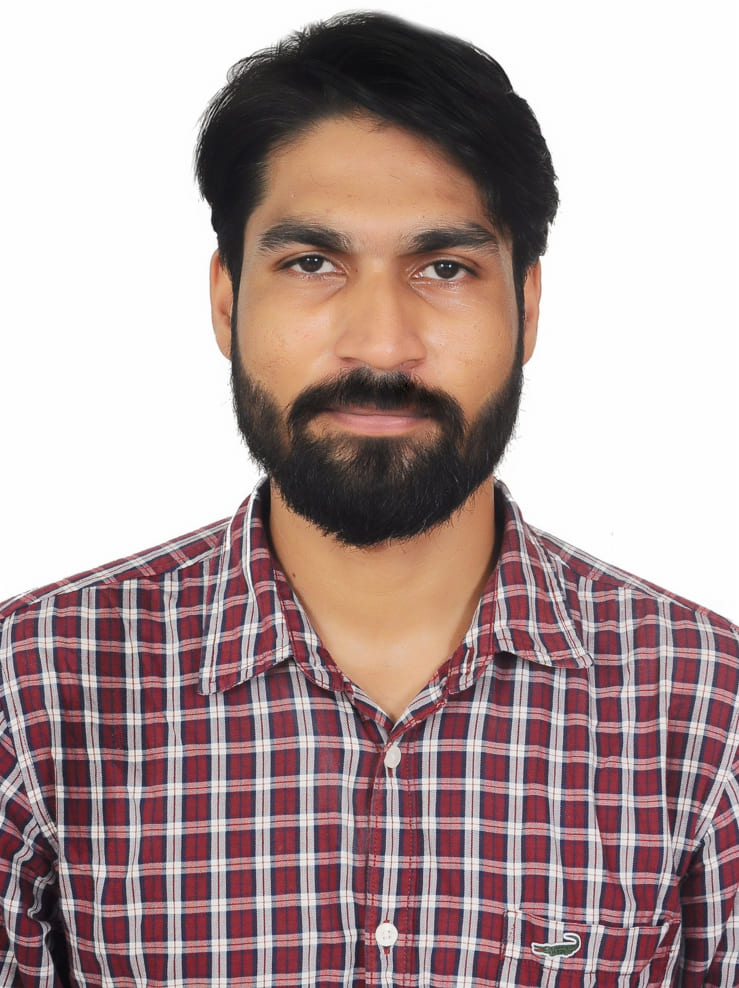}}]{Akhilesh Kumar Malviya} is currently pursing the M. Tech degree in Artificial Intelligence and Machine Learning from BITS Pilani, India. He holds an engineering degree in Electronics and communication Engineering.  He is a lead engineer in the department of Driving Software System (DSW), Valeo India. He has 7 years of experience in Computer vision, deep learning and Machine learning. He has authored 14 patents. His current research interests are multimodal learning, computer vision and image processing.
\end{IEEEbiography}
\vskip -1\baselineskip plus -1fil

\begin{IEEEbiography}[{\includegraphics[width=1in,height=1.25in,clip,keepaspectratio]{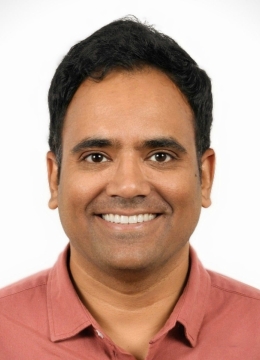}}]{Ganesh Sistu} serves as a Principal AI Architect at Valeo Ireland, where he leads global teams in applied research and product development, focusing on automated driving and parking perception. With a solid background of over 13 years in computer vision and machine learning, he has made notable contributions to the field, authoring over 30 publications in respected conferences including ICCV, WACV, ICRA, and GECCO. Balancing his industrial role, he also engages in academia as an Adjunct Assistant Professor at the University of Limerick, Ireland. His commitment extends to guiding emerging talents in AI, serving on the industry board for the Science Foundation Ireland Data Science PhD and the University of Limerick's National MSc in AI Programs.
\end{IEEEbiography}
\vskip -1\baselineskip plus -1fil

\begin{IEEEbiography}
[{\includegraphics[width=1in,height=1.25in,clip,keepaspectratio]{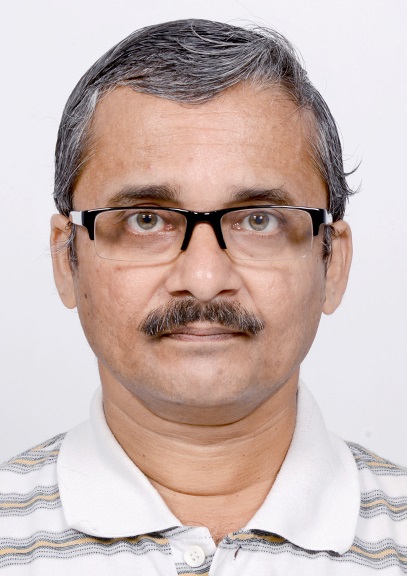}}]{Ujjwal Bhattacharya}
is currently a member of the faculty of the Computer Vision and Pattern Recognition Unit of Indian Statistical Institute situated in Kolkata. He joined his current Institute in 1991 as a Junior Research Fellow after obtaining his M.Sc. and M. Phil. degrees in Pure Mathematics from Calcutta University. 
In the past, he collaborated with a few  industries and research labs of India and abroad. 
In 1995, he received Young Scientist Award from the Indian Science Congress Association. Also, he received a few best paper awards from various groups. He is a senior member of the IEEE and a life member of IUPRAI, Indian unit of the IAPR. He has served as Program Committee member of various reputed International Conferences and Workshops. 
Also, he worked as a Co-Guest Editor of a few Special Issues of International Journals. His current research interests include machine learning, computer vision, image processing, document processing, handwriting recognition etc.
\end{IEEEbiography}
\vskip -1\baselineskip plus -1fil

\begin{IEEEbiography}
[{\includegraphics[width=1in,height=1.25in,clip,keepaspectratio]{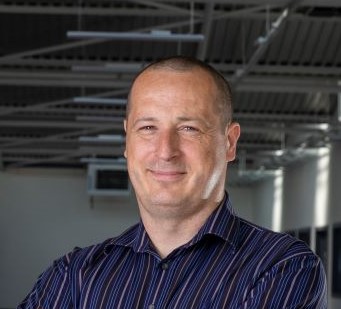}}]{Ciarán Eising} received the B.E. in Electronic and Computer Engineering and the Ph.D. degree from the National University of Ireland in 2003 and 2010, respectively. From 2009 to 2020, he worked as a Computer Vision Team Lead and an Architect at Valeo Vision Systems, where he also held the title of Senior Expert. In 2016, he was awarded the position of Adjunct Lecturer at the National University of Ireland, Galway. In 2020, he joined the University of Limerick as a Lecturer in Artificial Intelligence and Computer Vision. Ciarán is a Senior Member of the IEEE.
\end{IEEEbiography} 
\vskip -1\baselineskip plus -1fil